\documentclass[runningheads]{llncs}
\usepackage{graphicx}

\usepackage{tikz}
\usepackage{comment}
\usepackage{amsmath,amssymb} 
\usepackage{color}

\usepackage[accsupp]{axessibility}  

\usepackage{amsmath}
\usepackage{booktabs}
\usepackage{multirow}
\usepackage{orcidlink}
\usepackage{relsize}

\usepackage[capitalize]{cleveref}
\crefname{section}{Sec.}{Secs.}
\Crefname{section}{Section}{Sections}
\Crefname{table}{Table}{Tables}
\crefname{table}{Tab.}{Tabs.}

\def\R{\mathbb{R}}
\def\N{\mathbb{N}}

\DeclareMathOperator*{\argmax}{arg\,max}
\DeclareMathOperator*{\argmin}{arg\,min}
\DeclareMathOperator*{\mean}{mean}

\newtheorem{observation}{Observation}

\makeatletter
\renewcommand*\paragraph{
    \@startsection{paragraph}{4}{\z@}%
        {0.8ex \@plus1ex \@minus.2ex}%
        {-1em}%
        {\normalfont\normalsize\bfseries}}
\makeatother

\usepackage{multirow}

\usepackage{subcaption}
\usepackage{wrapfig}

\begin{document}
\pagestyle{headings}
\mainmatter
\def\ECCVSubNumber{3240}  

\title{Interpretable Image Classification with Differentiable Prototypes Assignment
}
\titlerunning{Interpret. img Class. with Differentiable Prototypes Assignment} 
\authorrunning{Rymarczyk et al.} 

\author{Dawid Rymarczyk$^{1, 2}\orcidlink{0000-0002-3406-6732x}$
\and
Łukasz Struski$^{1}$\orcidlink{0000-0003-4006-356X}
\and
Michał Górszczak$^{1}$\orcidlink{0000-0003-3695-0975}
\and \\
Koryna Lewandowska$^{3}$\orcidlink{0000-0003-4826-6361}
\and
Jacek Tabor$^{1}$\orcidlink{0000-0001-6652-7727}
\and
Bartosz Zieliński$^{1, 2}$\orcidlink{0000-0002-3063-3621}\\
}
\institute{Faculty of Mathematics and Computer Science, Jagiellonian University\\
\and
Ardigen SA\\
\and 
Department of Cognitive Neuroscience and Neuroergonomics, \\Institute of Applied Psychology, Jagiellonian University}

\maketitle

\begin{abstract}
Existing prototypical-based models address the black-box nature of deep learning. However, they are sub-optimal as they often assume separate prototypes for each class, require multi-step optimization, make decisions based on prototype absence (so-called negative reasoning process), and derive vague prototypes.
To address those shortcomings, we introduce ProtoPool, an interpretable prototype-based model with positive reasoning and three main novelties. Firstly, we reuse prototypes in classes, which significantly decreases their number. Secondly, we allow automatic, fully differentiable assignment of prototypes to classes, which substantially simplifies the training process. Finally, we propose a new focal similarity function that contrasts the prototype from the background and consequently concentrates on more salient visual features.
We show that ProtoPool obtains state-of-the-art accuracy on the CUB-200-2011 and the Stanford Cars datasets, substantially reducing the number of prototypes. We provide a theoretical analysis of the method and a user study to show that our prototypes capture more salient features than those obtained with competitive methods. We made the code available at \url{https://github.com/gmum/ProtoPool}.

\keywords{deep learning; interpretability; case-based reasoning}
\end{abstract}

\section{Introduction}
\label{sec:intro}

The broad application of deep learning in fields like medical diagnosis~\cite{afnan2021interpretable} and autonomous driving~\cite{wiegand2019drive}, together with current law requirements (such as GDPR in EU~\cite{kaminski2021right}), enforces models to explain the rationale behind their decisions. That is why explainers~\cite{basaj2021explaining,kim2018interpretability,lundberg2017unified,ribeiro2016should,selvaraju2017grad} and self-explainable~\cite{NEURIPS2018_3e9f0fc9,brendel2018approximating,zheng2019looking} models are developed to justify neural network predictions. Some of them are inspired by mechanisms used by humans to explain their decisions, like matching image parts with memorized prototypical features that an object can poses~\cite{chen2019looks,li2018deep,nauta2021neural,rymarczyk2021protopshare}.

Recently, a self-explainable model called Prototypical Part Network (ProtoPNet)~\cite{chen2019looks} was introduced, employing feature matching learning theory~\cite{rosch1975cognitive,rosch1973natural}. It focuses on crucial image parts and compares them with reference patterns (prototypical parts) assigned to classes. The comparison is based on a similarity metric between the image activation map and representations of prototypical parts (later called prototypes). The maximum value of similarity is pooled to the classification layer. As a result, ProtoPNet explains each prediction with a list of reference patterns and their similarity to the input image. Moreover, a global explanation can be obtained for each class by analyzing prototypical parts assigned to particular classes.

\begin{figure}[t]
    \centering
    \includegraphics[width=.75\columnwidth]{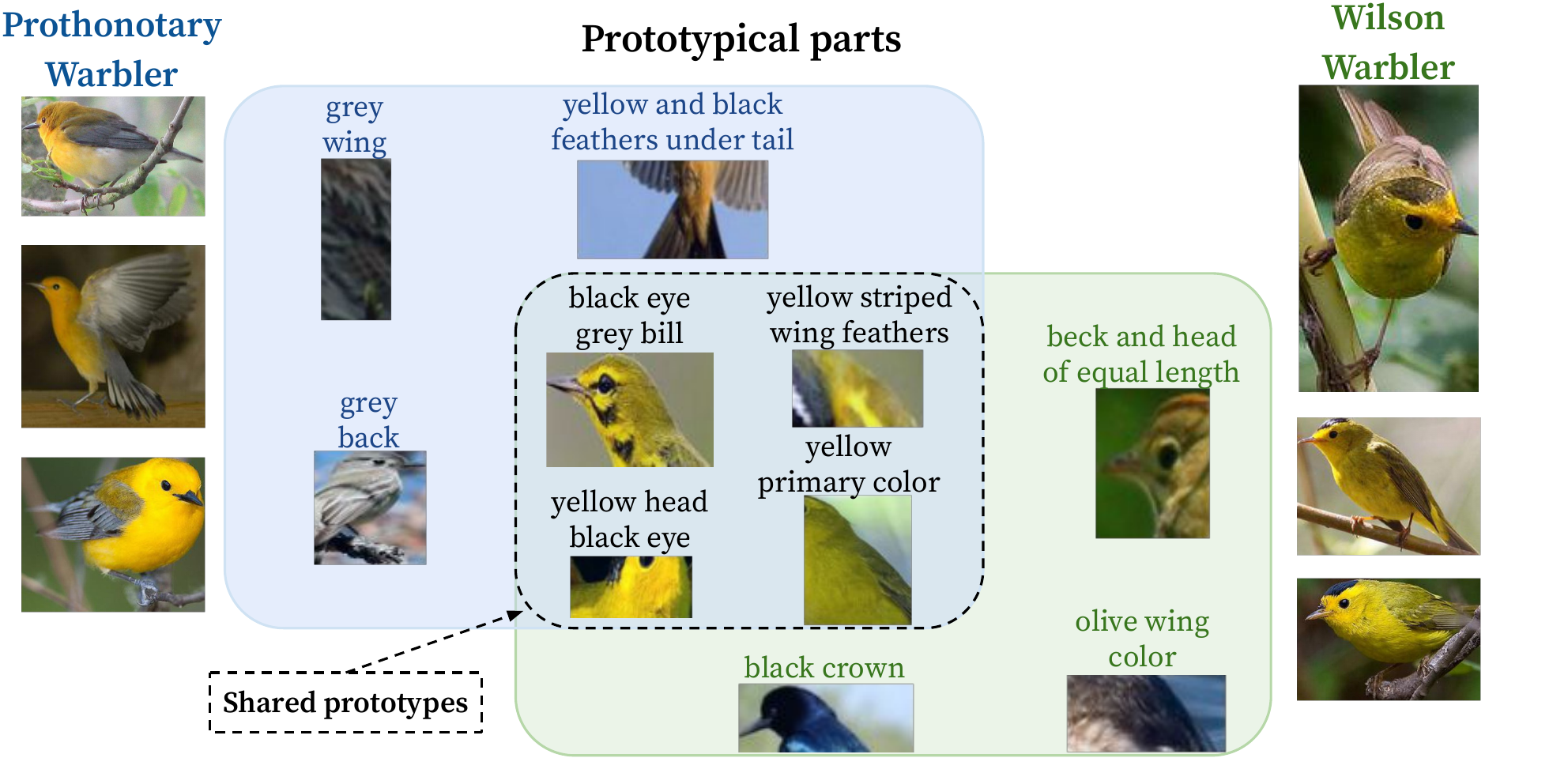}
    \caption{Automatically discovered prototypes\protect\footnotemark{} for two classes, \textit{Prothonotary Warbler} and \textit{Wilson Warbler} (each class represented by three images on left and right side). Three prototypical parts on the blue and green background are specific for a \textit{Prothonotary Warbler} and \textit{Wilson Warbler}, respectively (they correspond to heads and wings feathers). At the same time, four prototypes shared between those classes (related to yellow feathers) are presented in the intersection. Prototypes sharing reduces their amount, leads to a more interpretable model, and discovers classes similarities.}
    \label{fig:similar_classes}
\end{figure}
\footnotetext{Names of prototypical parts were generated based on the annotations from CUB-200-2011 dataset (see details in Supplementary Materials).}

However, ProtoPNet assumes that each class has its own separate set of prototypes, which is problematic because many visual features occur in many classes. For instance, both \textit{Prothonotary Warbler} and \textit{Wilson Warbler} have yellow as a primary color (see \Cref{fig:similar_classes}). Such limitation of ProtoPNet hinders the scalability because the number of prototypes grows linearly growing number of classes. Moreover, a large number of prototypes makes ProtoPNet hard to interpret by the users and results in many background prototypes~\cite{rymarczyk2021protopshare}.

To address these limitations, ProtoPShare~\cite{rymarczyk2021protopshare} and ProtoTree~\cite{nauta2021neural} were introduced. They share the prototypes between classes but suffer from other drawbacks. ProtoPShare requires previously trained ProtoPNet to perform the merge-pruning step, which extends the training time. At the same time, ProtoTree builds a decision tree and exploits the negative reasoning process that may result in explanations based only on prototype absence. 
For example, a model can predict a \textit{sparrow} because an image does not contain red feathers, a long beak, and wide wings. While this characteristic is true in the case of a \textit{sparrow}, it also matches many other species.

\begin{wrapfigure}{R}{0.4\textwidth}
\vspace{-3em}
\centering
    \centering
    \includegraphics[width=0.4\textwidth]{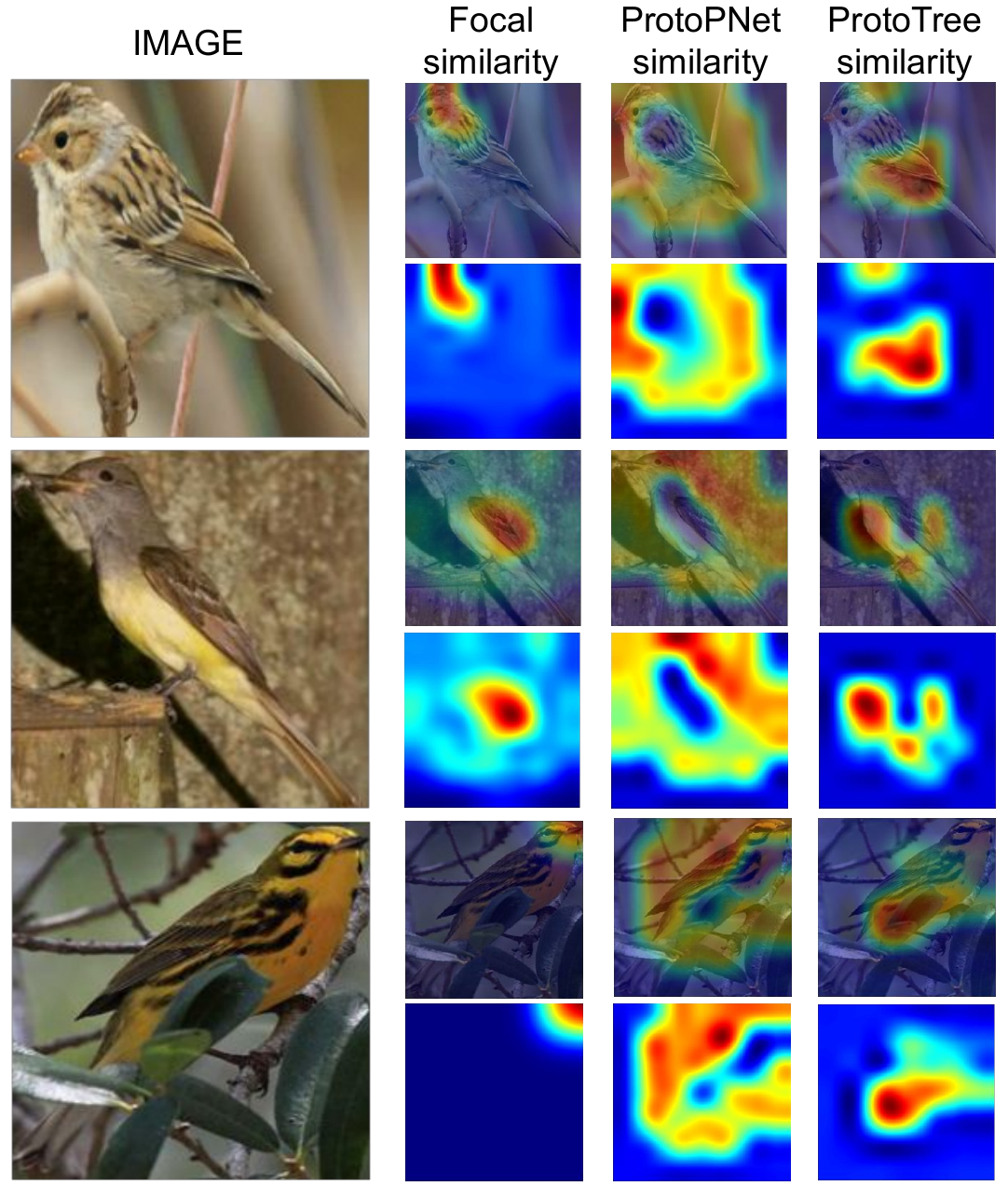}
    \caption{Focal similarity focuses the prototype on a salient visual feature. While the other similarity metrics are more distributed through the image, making the interpretation harder to comprehend. It is shown with three input images, the prototype activation map, and its overlay.}
    \label{fig:diff_sim_intro}
        \vspace{-2em}
\end{wrapfigure}

To deal with the above shortcomings, we introduce ProtoPool, a self-explainable prototype model for fine-grained images classification. ProtoPool introduces significantly novel mechanisms that substantially reduce the number of prototypes and obtain higher interpretability and easier training. Instead of using hard assignment of prototypes to classes, we implement the soft assignment represented by a distribution over the set of prototypes. This distribution is randomly initialized and binarized during training using the Gumbel-Softmax trick. Such a mechanism simplifies the training process by removing the pruning step required in ProtoPNet, ProtoPShare, and ProtoTree. The second novelty is a focal similarity function that focuses the model on the salient features. For this purpose, instead of maximizing the global activation, we widen the gap between the maximal and average similarity between the image activation map and prototypes (see \Cref{fig:similarity}). As a result, we reduce the number of prototypes and use the positive reasoning process on salient features, as presented in \Cref{fig:diff_sim_intro} and \Cref{fig.user_study}.

We confirm the effectiveness of ProtoPool with theoretical analysis and exhaustive experiments, showing that it achieves the highest accuracy among models with a reduced number of prototypes. What is more, we discuss interpretability, perform a user study, and discuss the cognitive aspects of the ProtoPool over existing methods.

The main achievements of the paper can be summarized as follows:
\begin{itemize}
  \setlength{\parskip}{0pt}
  \setlength{\itemsep}{2pt plus 2pt}
    \item We construct ProtoPool, a case-based self-explainable method that shares prototypes between data classes without any predefined concept dictionary.
    \item We introduce fully differentiable assignments of prototypes to classes, allowing the end-to-end training.
    \item We define a novel similarity function, called focal similarity, that focuses the model on the salient features.
    \item We increase interpretability by reducing prototypes number and providing explanations in a positive reasoning process.
\end{itemize}

\section{Related works}

Attempts to explain deep learning models can be divided into the post hoc and self-explainable~\cite{rudin2019stop} methods. The former approaches assume that the reasoning process is hidden in a black box model and a new explainer model has to be created to reveal it. Post hoc methods include a saliency map~\cite{marcos2019semantically,rebuffi2020there,selvaraju2017grad,selvaraju2019taking,simonyan2014deep} generating a heatmap of crucial image parts, or Concept Activation Vectors (CAV) explaining the internal network state as user-friendly concepts~\cite{chen2020concept,NEURIPS2019_77d2afcb,kim2018interpretability,pmlr-v119-koh20a,NEURIPS2020_ecb287ff}. Other methods provide counterfactual examples~\cite{abbasnejad2020counterfactual,goyal2019counterfactual,mothilal2020explaining,niu2021counterfactual,wang2020scout} or analyze the networks' reaction to the image perturbation~\cite{basaj2021explaining,fong2019understanding,fong2017interpretable,ribeiro2016should}. Post hoc methods are easy to implement because they do not interfere with the architecture, but they can produce biased and unreliable explanations~\cite{NEURIPS2018_294a8ed2}. That is why more focus is recently put on designing self-explainable models~\cite{NEURIPS2018_3e9f0fc9,brendel2018approximating} that make the decision process directly visible. Many interpretable solutions are based on the attention~\cite{liu2021visual,sundararajan2017axiomatic,xiao2015application,zheng2017learning,zheng2019looking,zhou2018interpretable} or exploit the activation space~\cite{Guidotti_Monreale_Matwin_Pedreschi_2020,puyol2020interpretable}, e.g. with adversarial autoencoder. However, most recent approaches built on an interpretable method introduced in~\cite{chen2019looks} (ProtoPNet) with a hidden layer of prototypes representing the activation patterns.

ProtoPNet inspired the design of many self-explainable models, such as TesNet ~\cite{wang2021interpretable} that constructs the latent space on a Grassman manifold without prototypes reduction. Other models like ProtoPShare~\cite{rymarczyk2021protopshare} and ProtoTree~\cite{nauta2021neural} reduce the number of prototypes used in the classification. The former introduces data-dependent merge-pruning that discovers prototypes of similar semantics and joins them. The latter uses a soft neural decision tree that may depend on the negative reasoning process. Alternative approaches organize the prototypes hierarchically~\cite{hase2019interpretable} to classify input at every level of a predefined taxonomy or transform prototypes from the latent space to data space~\cite{li2018deep}. Moreover, prototype-based solutions are widely adopted in various fields such as medical imaging~\cite{afnan2021interpretable,barnett2021iaia,kim2021xprotonet,singh2021these}, time-series analysis~\cite{gee2019explaining}, graphs classification~\cite{zhang2021protgnn}, and sequence learning~\cite{ming2019interpretable}.

\begin{figure}[t]
    \centering
    \includegraphics[width=.85\textwidth]{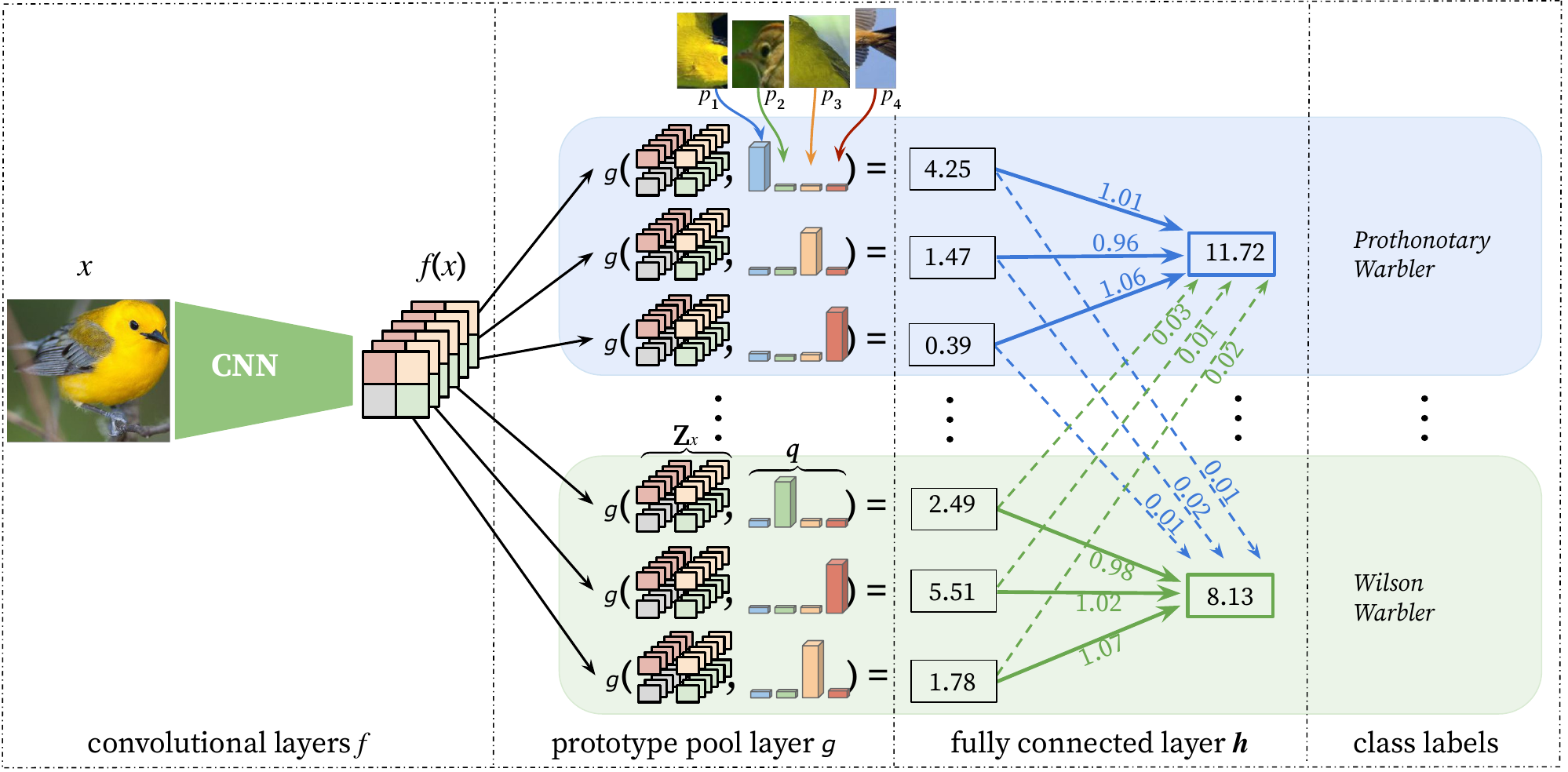}
    \caption{The architecture of our ProtoPool with a prototype pool layer $g$. Layer $g$ contains a pool of prototypes $p_1-p_4$ and three slots per class. Each slot is implemented as a distribution $q\in \mathbb{R}^4$ of prototypes from the pool, where successive values of $q$ correspond to the probability of assigning successive prototypes to the slot. In this example, $p_1$ and $p_2$ are assigned to the first slot of \textit{Prothonotary Warbler} and \textit{Wilson Warbler}, respectively. At the same time, the shared prototypes $p_3$ and $p_4$ are assigned to the second and third slots of both classes.}
    \label{fig:protopool}
\end{figure}

\section{ProtoPool}
\label{sec:method}

In this section, we describe the overall architecture of ProtoPool presented in \Cref{fig:protopool} and the main novelties of ProtoPool compared to the existing models, including the mechanism of assigning prototypes to slots and the focal similarity. Moreover, we provide a theoretical analysis of the approach.

\paragraph{Overall architecture}
The architecture of ProtoPool, shown in \Cref{fig:protopool}, is generally inspired by ProtoNet~\cite{chen2019looks}. It consists of convolutional layers $f$, a prototype pool layer $g$, and a fully connected layer $h$. Layer $g$ contains a pool of $M$ trainable prototypes $P=\{p_i\in \mathbb{R}^D\}_{i=1}^{M}$ and $K$ slots for each class. Each slot is implemented as a distribution $q_k\in \mathbb{R}^M$ of prototypes available in the pool, where successive values of $q_k$ correspond to the probability of assigning successive prototypes to slot $k$ ($\|q_k\|=1$). Layer $h$ is linear and initialized to enforce the positive reasoning process, i.e. weights between each class $c$ and its slots are initialized to $1$ while remaining weights of $h$ are set to $0$.

Given an input image $x \in X$, the convolutional layers first extract image representation $f(x)$ of shape $H\times W\times D$, where $H$ and $W$ are the height and width of representation obtained at the last convolutional layer for image $x$, and $D$ is the number of channels in this layer. Intuitively, $f(x)$ can be considered as a set of $H\cdot W$ vectors of dimension $D$, each corresponding to a specific location of the image (as presented in \Cref{fig:protopool}). For the clarity of description, we will denote this set as $Z_x=\{z_i\in f(x) : z_i\in \mathbb{R}^D, i=1,...,H\cdot W\}$. Then, the prototype pool layer is used on each $k$-th slot to compute the aggregated similarity $g_k=\sum_{i=1}^M q^i_k g_{p_i}$ between $Z_x$ and all prototypes considering the distribution $q_k$ of this slot, where $g_p$ is defined below. Finally, the similarity scores ($K$ values per class) are multiplied by the weight matrix $w_h$ in the fully connected layer $h$. This results in the output logits, further normalized using softmax to obtain a final prediction.

\paragraph{Focal similarity}
In ProtoPNet~\cite{chen2019looks} and other models using prototypical parts, the similarity of point $z$ to prototype $p$ is defined as\footnote{The following regularization is used to avoid numerical instability in the experiments: $g_p(z)=\log(\frac{\|z-p\|^2+1}{\|z-p\|^2+\varepsilon})$, with a small $\varepsilon>0$.}
$
\addtolength\abovedisplayskip{-0.5\baselineskip}%
\addtolength\belowdisplayskip{-0.5\baselineskip}%
g_p(z)=\log(1+\tfrac{1}{\|z-p\|^2}),
$
and the final activation of the prototype $p$ with respect to image $x$ is given by
$
\addtolength\abovedisplayskip{-0.5\baselineskip}%
\addtolength\belowdisplayskip{-0.5\baselineskip}%
g_p=\max_{z \in Z_x} g_p(z).
$
One can observe that such an approach has two possible disadvantages. First, high activation can be obtained when all the elements in $Z_x$ are similar to a prototype. It is undesirable because the prototypes can then concentrate on the background. The other negative aspect concerns the training process, as the gradient is passed only through the most active part of the image.

\begin{figure}[t]
    \centering
    \includegraphics[width=.7\columnwidth]{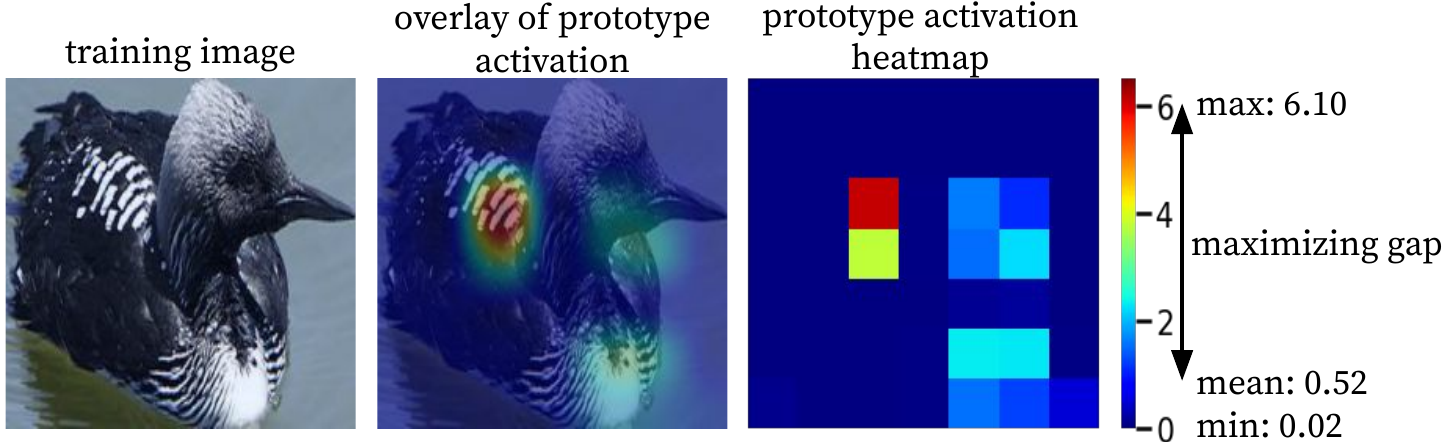}
    \caption{Our focal similarity limits high prototype activation to a narrow area (corresponding to white and black striped wings). It is obtained by widening the gap between the maximal and average activation (equal $6.10$ and $0.52$, respectively). As a result, our prototypes correspond to more salient features (according to our user studies described in Section~\ref{sec:interpretability}).}
    \label{fig:similarity}
\end{figure}

To prevent those behaviors, in ProtoPool, we introduce a novel focal similarity function that widens the gap between maximal and average activation
\begin{equation} \label{eq:contr}
\addtolength\abovedisplayskip{-0.5\baselineskip}%
\addtolength\belowdisplayskip{-0.5\baselineskip}%
g_p=\max_{z \in Z_x} g_p(z)-\mean_{z \in Z_x} g_p(z),
\end{equation}
as presented in \Cref{fig:similarity}.
The maximal activation of focal similarity is obtained if a prototype is similar to only a narrow area of the image $x$ (see \Cref{fig:diff_sim_intro}). Consequently, the constructed prototypes correspond to more salient features (according to our user studies described in Section~\ref{sec:interpretability}), and the gradient passes through all elements of $Z_x$.

\begin{figure}[t]
    \centering
    \includegraphics[width=.7\columnwidth]{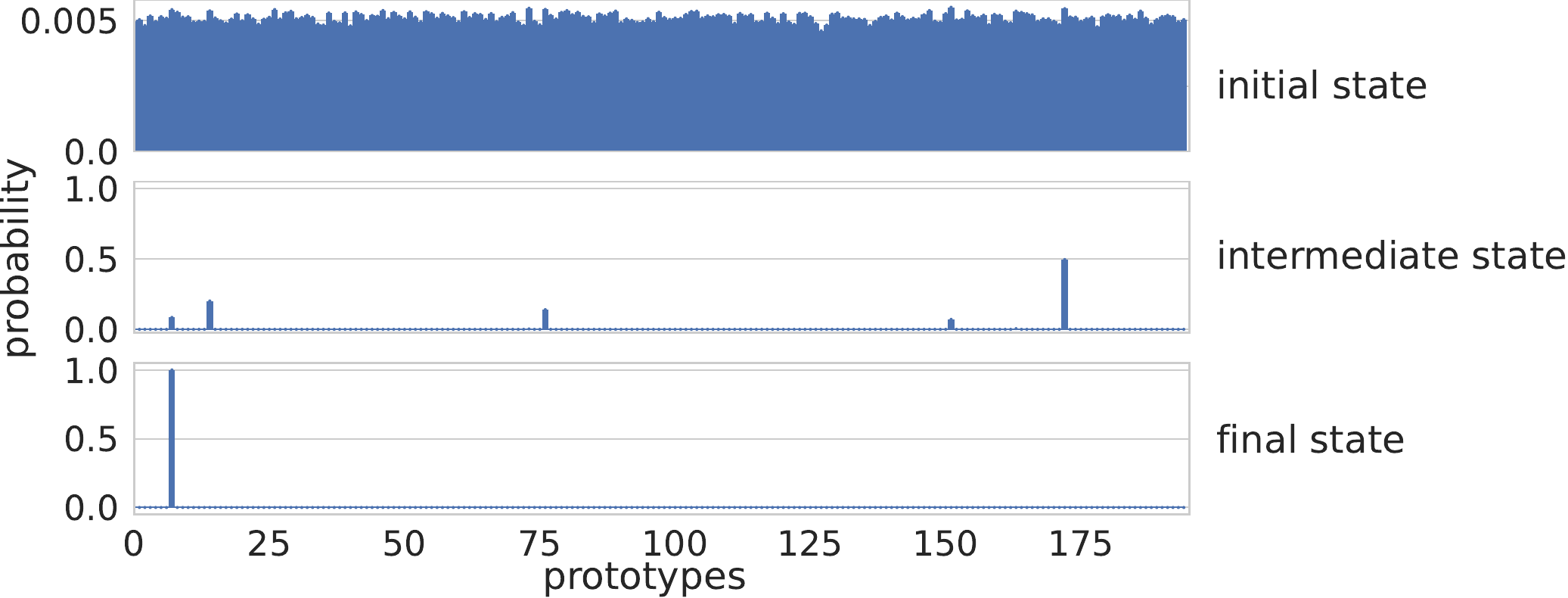}
    \caption{A sample distribution (slot) at the initial, middle, and final step of training. In the beginning, all prototypes are assigned with a probability of $0.005$. Then, the distribution binarizes, and finally, one prototype is assigned to this slot with a probability close to $1$.}
    \label{fig:q_process}
\end{figure}

\paragraph{Assigning one prototype per slot}
Previous prototypical methods use the hard predefined assignment of the prototypes to classes~\cite{chen2019looks,rymarczyk2021protopshare,wang2021interpretable} or nodes of a tree~\cite{nauta2021neural}. Therefore, no gradient propagation is needed to model the prototypes assignment. In contrast, our ProtoPool employs a soft assignment based on prototypes distributions to use prototypes from the pool optimally.
To generate prototype distribution $q$, one could apply softmax on the vector of size $\mathbb{R}^M$. However, this could result in assigning many prototypes to one slot and consequently could decrease the interpretability. Therefore, to obtain distributions with exactly one probability close to $1$, we require a differentiable $\argmax$ function. A perfect match, in this case is the Gumbel-Softmax estimator~\cite{jang2016categorical}, where for $q=(q^1,\ldots,q^M)\in\R^M$ and $\tau\in(0, \infty)$
\[
\addtolength\abovedisplayskip{-0.2\baselineskip}%
\addtolength\belowdisplayskip{-0.2\baselineskip}%
\mathrm{Gumbel\textnormal{-}softmax}(q, \tau) = (y^1, \ldots, y^M)\in\R^M,
\]
where 
\(
y^i = \tfrac{\exp\left((q^i + \eta_i) / \tau\right)}{\sum^M_{m=1}\exp\left((q^m + \eta_m) / \tau\right)}
\)
and $\eta_m$ for $m\in 1, .., M$ are samples drawn from standard Gumbel distribution.
The Gumbel-Softmax distribution interpolates between continuous categorical densities and discrete one-hot-encoded categorical distributions, approaching the latter for low temperatures $\tau\in[0.1, 0.5]$ (see \Cref{fig:q_process}).

\paragraph{Slots orthogonality}
Without any additional constraints, the same prototype could be assigned to many slots of one class, wasting the capacity of the prototype pool layer and consequently returning poor results. Therefore, we extend the loss function with
\begin{equation}\label{eq.loss_orth}
 \addtolength\abovedisplayskip{-0.5\baselineskip}%
 \addtolength\belowdisplayskip{-0.5\baselineskip}%
    \mathcal{L}_{\small orth} = \mathsmaller{\sum}\limits^{\mathsmaller{K}}_{\mathsmaller{i< j}}\tfrac{\langle q_i, q_j\rangle}{\|q_i\|_2\cdot \|q_j\|_2},
\end{equation}
where $q_1,..,q_K$ are the distributions of a particular class. As a result, successive slots of a class are assigned to different prototypes.


\paragraph{Prototypes projection}
Prototypes projection is a step in the training process that allows prototypes visualization. It replaces each abstract prototype learned by the model with the representation of the nearest training patch. For prototype $p$, it can be expressed by the following formula
\begin{equation}
 \addtolength\abovedisplayskip{-0.5\baselineskip}%
 \addtolength\belowdisplayskip{-0.5\baselineskip}%
\begin{gathered}
p \leftarrow \argmin_{z \in Z_{C}} \lVert{z-p}\rVert_2,
\end{gathered}
\end{equation}
where $Z_{C} = \{z: z \in Z_x \textnormal{ for all } (x,y): y \in C\}$. In contrast to~\cite{chen2019looks}, set $C$ is not a single class but the set of classes assigned to prototype $p$.

\paragraph{Theoretical analysis}
Here, we theoretically analyze why ProtoPool assigns one prototype per slot and why each prototype does not repeat in a class. For this purpose, we provide two observations.

\begin{observation}
Let $q\in [0,1]^M$, $\sum q_i=1$ be a distribution (slot) of a particular class. Then, the limit of $\mathrm{Gumbel\textnormal{-}softmax}(q, \tau)$, as $\tau$ approaches zero, is the canonical vector $e_i\in\R^M$, i.e. for $q$ there exists $i=1,..,M$ such that $\lim\limits_{\tau\to 0} \mathrm{Gumbel\textnormal{-}softmax}(q, \tau) = e_i$.
\end{observation}

\noindent The temperature parameter $\tau>0$ controls how closely the new samples approximate discrete one-hot vectors (the canonical vector). From paper~\cite{jang2016categorical} we know that as $\tau\to 0$, the $\mathrm{softmax}$ computation smoothly approaches the $\argmax$, and the sample vectors approach one-hot $q$ distribution (see \Cref{fig:q_process}).

\begin{observation}
Let $K\in\N$ and $q_1,..,q_K$ be the distributions (slots) of a particular class. If $\mathcal{L}_{\small orth}$ defined in Eq.~\eqref{eq.loss_orth} is zero, then each prototype from a pool is assigned to only one slot of the class.
\end{observation}

\noindent It follows the fact that $\mathcal{L}_{\small orth} = 0$ only if $\langle q_i, q_j\rangle = 0$ for all $i<j\leq K$, i.e. only if $q_i, q_j$ have non-zero values for different prototypes.

\begin{table}[t]
  \centering
  \small
  \caption{Comparison of ProtoPool with other prototypical methods trained on the CUB-200-2011 and Stanford Cars datasets, which considers a various number of prototypes and types of convolutional layers $f$. In the case of the CUB-200-2011 dataset, ProtoPool achieves the highest accuracy than other models, even those containing ten times more prototypes. Moreover, the ensemble of three ProtoPools surpasses the ensemble of five TesNets with 17 times more prototypes. On the other hand, in the case of Stanford Cars, ProtoPool achieves competitive results with significantly fewer prototypes. Please note that the results are first sorted by backbone network and then by the number of prototypes, R stands for ResNet, iN means pretrained on iNaturalist, and Ex is an ensemble of three or five models.}
  \label{tab:CUB_acc}

\fontsize{8}{10}\selectfont
\begin{tabular}{l@{}c@{}c@{\;\;}l@{}}
    \toprule
    \multicolumn{4}{c}{\textbf{CUB-200-2011}} \\
    \midrule
    \multicolumn{1}{l}{Model} & Arch. & Proto. \# & Acc [\%] \\
    \cmidrule[0.5pt]{1-4}
    ProtoPool (ours) & \multirow{4}{*}{R34}  & 202 & $80.3\!\pm\!0.2$ \\
    ProtoPShare~\cite{rymarczyk2021protopshare} &  & $400$ & $74.7$ \\
    ProtoPNet~\cite{chen2019looks} & & $1655$ & $79.5$\\
    TesNet~\cite{wang2021interpretable} &  & $2000$ & $82.7\!\pm\!0.2$\\
    \cmidrule{1-4}
    ProtoPool (ours) & \multirow{4}{*}{R152} & 202 & $81.5\!\pm\!0.1$ \\
    ProtoPShare~\cite{rymarczyk2021protopshare} &  & $1000$ & $73.6$ \\
    ProtoPNet~\cite{chen2019looks} &  & $1734$ & $78.6$\\
    TesNet~\cite{wang2021interpretable} &  & $2000$ & $82.8\!\pm\!0.2$\\
    \cmidrule{1-4}
    ProtoPool (ours) & \multirow{2}{*}{iNR50} & $202$ & $85.5\!\pm\!0.1$\\
    ProtoTree~\cite{nauta2021neural} &  & $202$ & $82.2\!\pm\!0.7$\\
    \cmidrule{1-4}
    ProtoPool (ours) & \multirow{2}{*}{Ex3} & $202\!\!\times\!\!3$ & $87.5$\\
    ProtoTree~\cite{nauta2021neural} &  & $202\!\!\times\!\!3$ & $86.6$\\
    \cmidrule{1-4}
    ProtoPool (ours) & \multirow{4}{*}{Ex5} & $202\!\!\times\!\!5$ & $\mathbf{87.6}$\\
    ProtoTree~\cite{nauta2021neural} &  & $202\!\!\times\!\!5$ & $87.2$\\
    ProtoPNet~\cite{chen2019looks} &  & $2000\!\!\times\!\!5$ & $84.8$\\
    TesNet~\cite{wang2021interpretable} &  & $2000\!\!\times\!\!5$ & $86.2$\\
    \cmidrule[0.8px]{1-4}
\end{tabular}
~
\begin{tabular}{l@{}c@{\;\;}c@{\;\;}l@{\;}}
    \toprule
    \multicolumn{4}{c}{\textbf{Stanford Cars}} \\
    \midrule
    \multicolumn{1}{l}{Model} & Arch. & Proto. \# & Acc [\%] \\
    \cmidrule[0.5pt]{1-4}
    ProtoPool (ours) & \multirow{4}{*}{R34} & $195$ & $89.3\!\pm\!0.1$ \\
    ProtoPShare~\cite{rymarczyk2021protopshare} && $480$ & $86.4$ \\
    ProtoPNet~\cite{chen2019looks} && $1960$ & $86.1\!\pm\!0.2$\\
    TesNet~\cite{wang2021interpretable} && $1960$ & $92.6\!\pm\!0.3$\\
    \cmidrule{1-4}
    ProtoPool (ours) & \multirow{2}{*}{R50} & $195$ & $88.9\!\pm\!0.1$\\
    ProtoTree~\cite{nauta2021neural} &  & $195$ & $86.6\!\pm\!0.2$\\
    \cmidrule{1-4}
    ProtoPool (ours) & \multirow{2}{*}{Ex3} & $195\!\!\times\!\!3$ & $91.1$\\
    ProtoTree~\cite{nauta2021neural} &  & $195\!\!\times\!\!3$ & $90.5$\\
    \cmidrule{1-4}
    ProtoPool (ours) & \multirow{4}{*}{Ex5} & $195\!\!\times\!\!5$ & $91.6$\\
    ProtoTree~\cite{nauta2021neural} &  & $195\!\!\times\!\!5$ & $91.5$\\
    ProtoPNet~\cite{chen2019looks} &  & $1960\!\!\times\!\!5$ & $91.4$\\
    TesNet~\cite{wang2021interpretable} &  & $1960\!\!\times\!\!5$ & $\mathbf{93.1}$\\
    \cmidrule[0.8px]{1-4}
    &&& \\
    &&& \\
    &&& \\
    &&& \\[-4pt]
    &&& 
  \end{tabular}
\end{table}

\section{Experiments}

We train our model on CUB-200-2011~\cite{wah2011caltech} and Stanford Cars~\cite{krause20133d} datasets to classify 200 bird species and 196 car models, respectively. As the convolutional layers $f$ of the model, we take ResNet-34, ResNet-50, ResNet-121~\cite{he2016deep}, DenseNet-121, and DenseNet-161~\cite{huang2017densely} without the last layer, pretrained on ImageNet~\cite{deng2009imagenet}. The one exception is ResNet-50 used with CUB-200-2011 dataset, which we pretrain on iNaturalist2017~\cite{van2018inaturalist} for fair comparison with ProtoTree model~\cite{nauta2021neural}. In the testing scenario, we make the prototype assignment hard, i.e. we set all values of a distribution $q$ higher than $0.5$ to $1$, and the remaining values to $0$ otherwise. We set the number of prototypes assigned to each class to be at most $10$ and use the pool of $202$ and $195$ prototypical parts for CUB-200-2011 and Stanford Cars, respectively. Details on experimental setup and results for other backbone networks are provided in the Supplementary Materials.

\paragraph{Comparison with other prototypical models}
In \Cref{tab:CUB_acc} we compare the efficiency of our ProtoPool with other models based on prototypical parts. We report the mean accuracy and standard error of the mean for $5$ repetitions. Additionally, we present the number of prototypes used by the models, and we use this parameter to sort the results. We compare ProtoPool with ProtoPNet~\cite{chen2019looks}, ProtoPShare~\cite{rymarczyk2021protopshare}, ProtoTree~\cite{nauta2021neural}, and TesNet~\cite{wang2021interpretable}.

One can observe that ProtoPool achieves the highest accuracy for the CUB-200-2011 dataset, surpassing even the models with a much larger number of prototypical parts (TesNet and ProtoPNet). For Stanford Cars, our model still performs better than other models with a similarly low number of prototypes, like ProtoTree and ProtoPShare, and slightly worse than TesNet, which uses ten times more prototypes. The higher accuracy of the latter might be caused by prototype orthogonality enforced in training. Overall, our method achieves competitive results with significantly fewer prototypes. However, ensemble ProtoPool or TesNet should be used if higher accuracy is preferred at the expense of interpretability.

\begin{figure}[t]

\begin{minipage}{.48\textwidth}
    \centering
    \includegraphics[width=\textwidth]{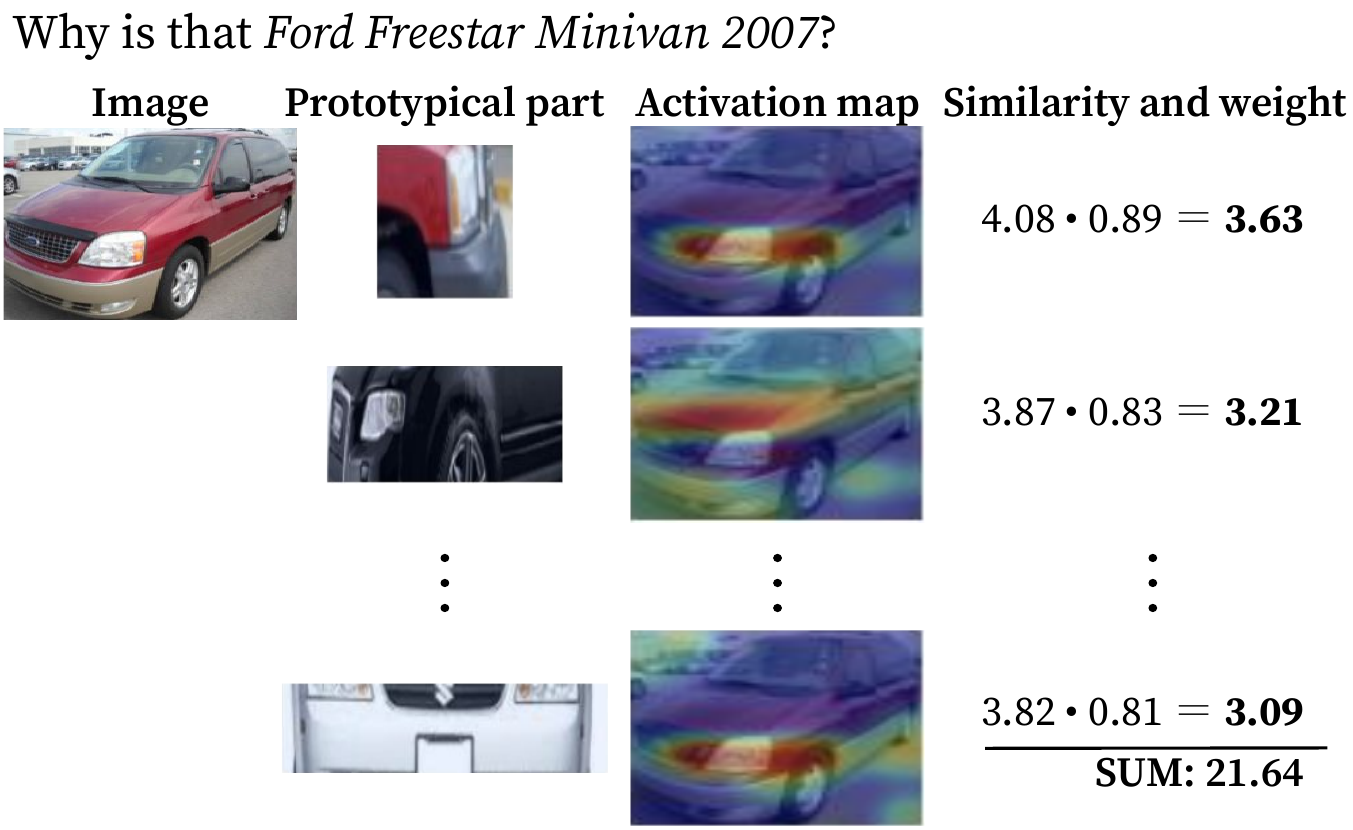}
    \caption{Sample explanation of \textit{Ford Freestar Minivan 2007} predictions. Except for an image, we present a few prototypical parts of this class, their activation maps, similarity function values, and the last layer weights. Moreover, we provide the sum of the similarities multiplied by the weights. ProtoPool returns the class with the largest sum as a prediction.}
    \label{fig:explanation}
\end{minipage}
~
\begin{minipage}{.48\textwidth}
    \centering
    \includegraphics[width=\textwidth]{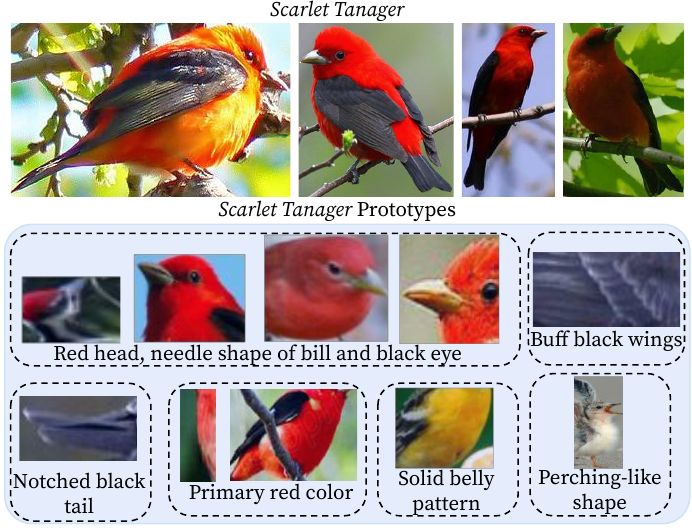}
    \caption{Samples of \textit{Scarlet Tanager} and prototypical parts assigned to this class by our ProtoPool model. Prototypes correspond, among others, to primary red color of feathers, black eye, perching-like shape, black notched tail, and black buff wings\protect\footnotemark{}.}
    \label{fig:global}
\end{minipage}
\end{figure}

\section{Interpretability}
\label{sec:interpretability}

In this section, we analyze the interpretability of the ProtoPool model. Firstly, we show that our model can be used for local and global explanations. Then, we discuss the differences between ProtoPool and other prototypical approaches, and investigate its stability. Then, we perform a user study on the similarity functions used by the ProtoPNet, ProtoTree, and ProtoPool to assess the saliency of the obtained prototypes. Lastly, we consider ProtoPool from the cognitive psychology perspective.

\begin{figure}[t]
    \centering
    \includegraphics[width=0.95\textwidth]{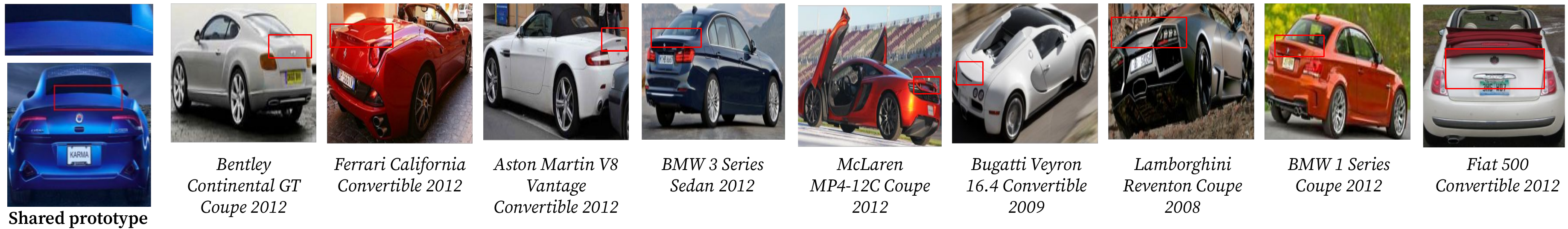}
    \caption{Sample prototype of a \textit{convex tailgate} (left top corner) shared by nine classes. Most of the classes correspond to luxury cars, but some exceptions exist, such as \textit{Fiat 500}.}
    \label{fig:shared_one}
\end{figure}

\paragraph{Local and global interpretations}
Except for local explanations that are similar to those provided by the existing methods (see \Cref{fig:explanation}), ProtoPool can provide a global characteristic of a class. It is presented in \Cref{fig:global}, where we show the prototypical parts of \textit{Scarlet Tanager} that correspond to the visual features of this species, such as red feathers, a puffy belly, and a short beak.
Moreover, similarly to ProtoPShare, ProtoPool shares the prototypical parts between data classes. Therefore, it can describe the relations between classes relying only on the positive reasoning process, as presented in \Cref{fig:similar_classes} (in contrast, ProtoTree also uses negative reasoning). In \Cref{fig:shared_one}, we further provide visualization of the prototypical part shared by nine classes. More examples are provided in Supplementary Materials.

\begin{table}[t]
  \caption{Characteristics of prototypical methods for fine-grained image classification that considers the number of prototypes, reasoning type, and prototype sharing between classes. ProtoPool uses 10\% of ProtoPNet's prototypes but only with positive reasoning. It shares the prototypes between classes but, in contrast to ProtoPShare, is trained in an end-to-end, fully differentiable manner. Please notice that 100\% corresponds to $2000$ and $1960$ of prototypes for CUB-200-2011 and Stanford Cars datasets, respectively.}
  \centering
  \fontsize{8}{10}\selectfont
  \begin{tabular}{@{}l@{\;\;}@{\;}c@{\quad}c@{\quad}c@{\quad}c@{\quad}c@{\;}}
    \toprule
    \textbf{Model} & ProtoPool & ProtoTree & ProtoPShare & ProtoPNet & TesNet \\
    \midrule
    \textbf{Portion of prototypes} & $\sim$10\% & $\sim$10\% & [20\%;50\%] & 100\% & 100\% \\
    \midrule
    \textbf{Reasoning type} & $+$ & $+/-$ & $+$ & $+$ & $+$ \\
    \midrule
    \textbf{Prototype sharing} & direct & indirect & direct & none & none \\
    \bottomrule
  \end{tabular}
  
  \label{tab:interpretability}
\end{table}

\paragraph{Differences between prototypical methods}
In \Cref{tab:interpretability}, we compare the characteristics of various prototypical-based methods. Firstly, ProtoPool and ProtoTree utilize fewer prototypical parts than ProtoPNet and TesNet (around 10\%). ProtoPShare also uses fewer prototypes (up to 20\%), but it requires a trained ProtoPNet model before performing merge-pruning. Regarding class similarity, it is directly obtained from ProtoPool slots, in contrast to ProtoTree, which requires traversing through the decision tree. Moreover, ProtoPNet and TesNet have no mechanism to detect inter-class similarities. Finally, ProtoTree depends, among others, on \textit{negative} reasoning process, while in the case of ProtoPool, it relies only on the positive reasoning process, which is a desirable feature according to~\cite{chen2019looks}.

\begin{figure}[t]
    \centering
    \includegraphics[width=0.75\textwidth]{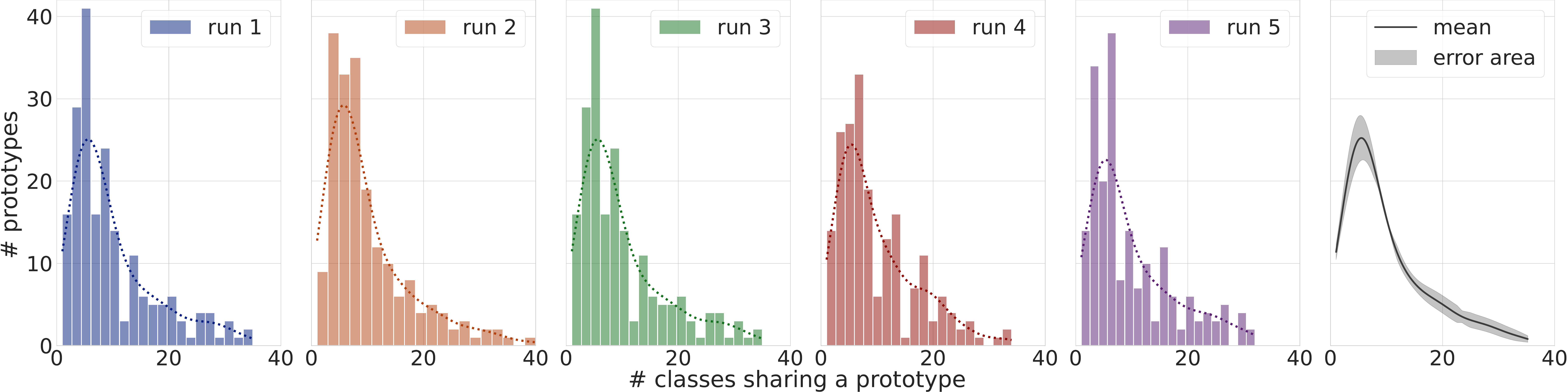}
    \caption{Distribution presenting how many prototypes are shared by the specific number of classes (an estimation plot is represented with a dashed line). Each color corresponds to a single ProtoPool training on Stanford Cars dataset with ResNet50 as a backbone network. The right plot corresponds to the mean and standard deviation for five training runs. One can observe that the distribution behaves stable between runs.}
    \label{fig:variability}
\end{figure}

\footnotetext{Names of prototypical parts were generated based on the annotations from CUB-200-2011 dataset (see details in Supplementary Materials).}
\paragraph{Stability of shared prototypes}
The natural question that appears when analyzing the assignment of the prototypes is: \textit{Does the similarity between two classes hold for many runs of ProtoPool training?}
To analyze this behavior, in \Cref{fig:variability} we show five distributions for five training runs. They present how many prototypes are shared by the specific number of classes. One can observe that difference between runs is negligible. In all runs, most prototypes are shared by five classes, but there exist prototypes shared by more than thirty classes. Moreover, on average, a prototype is shared by $2.73 \pm 0.51$ classes. A sample inter-class similarity graph is presented in the Supplementary Materials.

\paragraph{User study on focal similarity}\label{user}
To validate if using focal similarity results in more salient prototypical parts, we performed a user study where we asked the participants to answer the question: \textit{``How salient is the feature pointed out by the AI system?''}. The task was to assign a score from 1 to 5 where 1 meant \textit{``Least salient''} and 5 meant \textit{``Most salient''}.
Images were generated using prototypes obtained for ProtoPool with ProtoPNets similarity or with focal similarity and from a trained ProtoTree\footnote{ProtoTree was trained using code from~\url{https://github.com/M-Nauta/ProtoTree} and obtained accuracy similar to~\cite{nauta2021neural}. For ProtoPNet similarity, we used code from~\url{https://github.com/cfchen-duke/ProtoPNet}.}.
To perform the user study, we used Amazon Mechanical Turk (AMT) system\footnote{\url{https://www.mturk.com}}. To assure the reliability of the answers, we required the users to be masters according to AMT. $40$ workers participated in our study and answered $60$ questions ($30$ per dataset) presented in a random order, which resulted in $2400$ answers. Each question contained an original training image and the same image with overlayed activation map, as presented in \Cref{fig:diff_sim_intro}.


\begin{wrapfigure}{R}{0.47\textwidth}
    \vspace{-2em}
    \centering
    \includegraphics[width=0.45\textwidth]{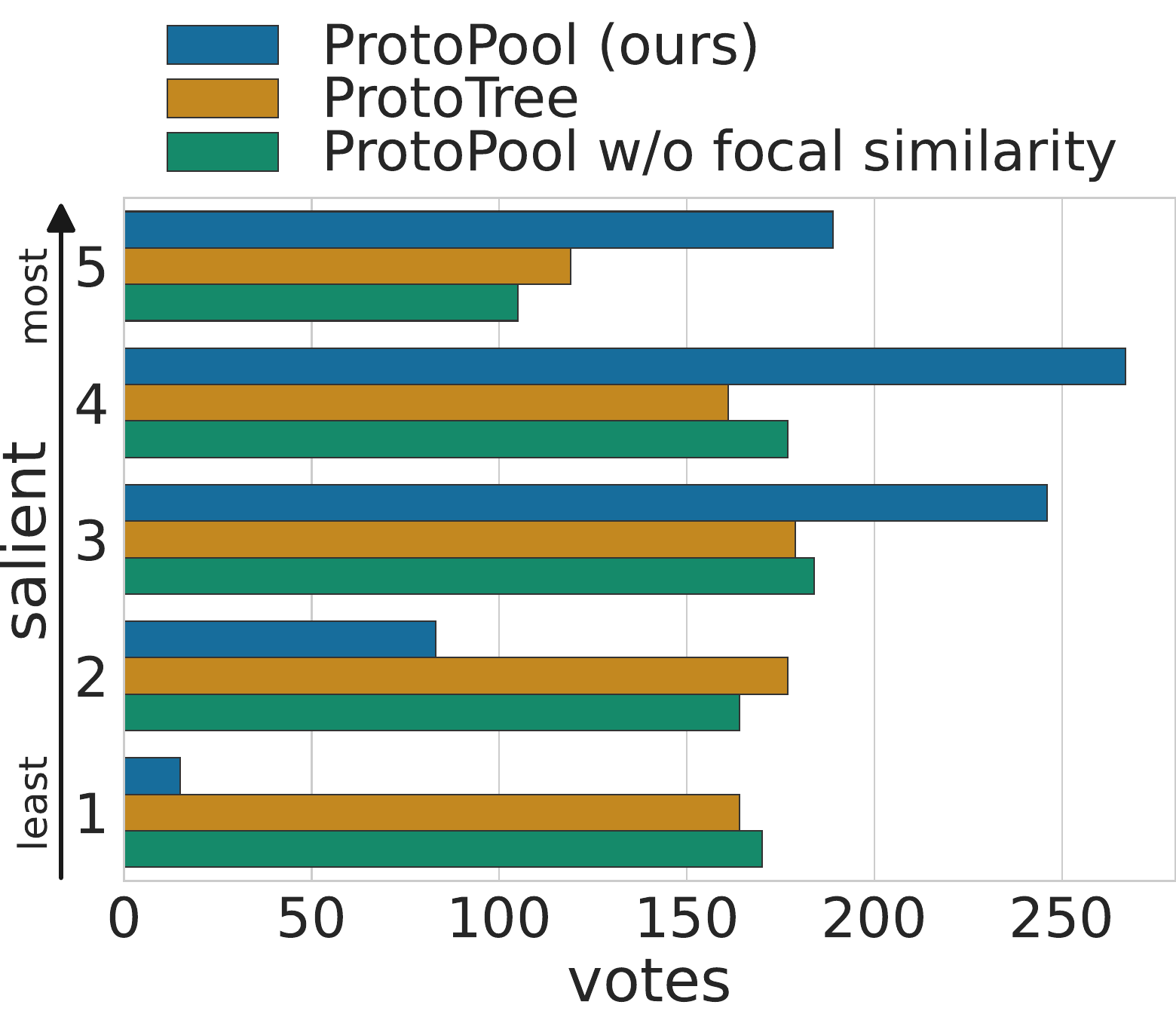}
    \caption{Distribution of scores from user study on prototypes obtained for ProtoPool without and with focal similarity and for ProtoTree. One can observe that ProtoPool with focal similarity generates more salient prototypes than the other models.}
    \label{fig.user_study}
    \vspace{-3em}
\end{wrapfigure}
Results presented in \Cref{fig.user_study} show that ProtoPool obtains mostly scores from 3 to 5, while other methods often obtain lower scores. We obtained a mean value of scores equal to $3.66$, $2.87$, and $2.85$ for ProtoPool, ProtoTree, and ProtoPool without focal similarity, respectively. Hence, we conclude that ProtoPool with focal similarity generated more salient prototypes than the reference models, including ProtoTree. See Supplementary Materials for more information about a user study, detailed results, and a sample questionnaire.

\paragraph{ProtoPool in the context of cognitive psychology}
ProtoPool can be described in terms of parallel or simultaneous information processing, while ProtoTree may be characterized by serial or successive processing, which takes more time~\cite{kesner1973neural,luria1973origin,neisser1967cognitive}.
More specifically, human cognition is marked with the speed-accuracy trade-off. Depending on the perceptual situation and the goal of a task, the human mind can apply a categorization process (simultaneous or successive) that is the most appropriate in a given context, i.e. the fastest or the most accurate. Both models have their advantages. However, ProtoTree has a specific shortcoming because it allows for a categorization process to rely on an absence of features. In other words, an object characterized by none of the enlisted features is labeled as a member of a specific category. This type of reasoning is useful when the amount of information to be processed (i.e. number of features and categories) is fixed and relatively small. However, the time of object categorization profoundly elongates if the number of categories (and therefore the number of features to be crossed out) is high. Also, the chance of miscategorizing completely new information is increased. 

\section{Ablation study}

In this section, we analyze how the novel architectural choices, the prototype projection, and the number of prototypes influence the model performance.

\begin{table}[tp]
  \centering
  \caption{The influence of prototype projection on ProtoPool performance for CUB-200-2011 and Stanford Cars datasets is negligible. Note that for CUB-200-2011, we used ResNet50 pretrained on iNaturalist.}
  \label{tab:push}
    \fontsize{9}{10}\selectfont
  \begin{tabular}{@{\;}c@{\qquad}c@{\;}c@{\qquad}c@{\;}c@{\;}}
  \toprule
    & \multicolumn{2}{c}{\textbf{CUB-200-2011}} & \multicolumn{2}{c}{\textbf{Stanford Cars}} \\
    \midrule
    Architecture & Acc [\%] before & Acc  [\%] after & Acc [\%] before & Acc  [\%] after \\
    \midrule
    ResNet34 & $80.8\!\pm\!0.2$ & $80.3\!\pm\!0.2$ & $89.1\!\pm\!0.2$ & $89.3\!\pm\!0.1$ \\
    ResNet50 & $85.9\!\pm\!0.1$ & $85.5\!\pm\!0.1$ & $88.4\!\pm\!0.1$ &  $88.9\!\pm\!0.1$ \\
    ResNet152 & $81.2\!\pm\!0.2$ & $81.5\!\pm\!0.1$ & --- & --- \\
    \bottomrule
  \end{tabular}
\end{table}

\paragraph{Influence of the novel architectural choices} 
Additionally, we analyze the influence of the novel components we introduce on the final results. For this purpose, we train ProtoPool without orthogonalization loss, with softmax instead of Gumbel-Softmax trick, and with similarity from ProtoPNet instead of focal similarity. Results are presented in \Cref{tab:without} and in Supplementary Materials. We observe that the Gumbel-Softmax trick has a significant influence on the model performance, especially for the Stanford Cars dataset, probably due to lower inter-class similarity than in CUB-200-2011 dataset~\cite{nauta2021neural}. On the other hand, the focal similarity does not influence model accuracy, although as presented in \Cref{user}, it has a positive impact on the interpretability. When it comes to orthogonality, it slightly increases the model accuracy by forcing diversity in slots of each class. Finally, the mix of the proposed mechanisms gets the best results.

\begin{table}[tp]
\caption{The influence of novel architectural choices on ProtoPool performance for CUB-200-2011 and Stanford Cars datasets is significant. We consider training without orthogonalization loss, with softmax instead of Gumbel-Softmax, and with similarity from ProtoPNet instead of focal similarity. One can observe that the mix of the proposed mechanisms (i.e. ProtoPool) obtains the best accuracy.}
  \centering
  \footnotesize
    \begin{tabular}{@{\;}lc@{\quad\quad}c}
    \toprule
    & \textbf{CUB-200-2011} & \textbf{Stanford Cars} \\
    \midrule
    Model & Acc [\%] & Acc [\%] \\
    \midrule
    ProtoPool  & $\mathbf{85.5}$ & $\mathbf{88.9}$\\
    w/o $\mathcal{L}_{\small orth}$ & $82.4 $ & $86.8$ \\
    w/o Gumbel-Softmax trick & $80.3 $ & $64.5$ \\
    w/o Gumbel-Softmax trick and $\mathcal{L}_{\small orth}$ & $65.1 $ & $30.8 $ \\
    w/o focal similarity & $85.3 $ & $88.8$ \\
    \bottomrule
  \end{tabular}
  
  \label{tab:without}
\end{table}

\begin{figure}[ht]
     \centering
     \begin{subfigure}[t]{0.49\textwidth}
         \includegraphics[width=\textwidth]{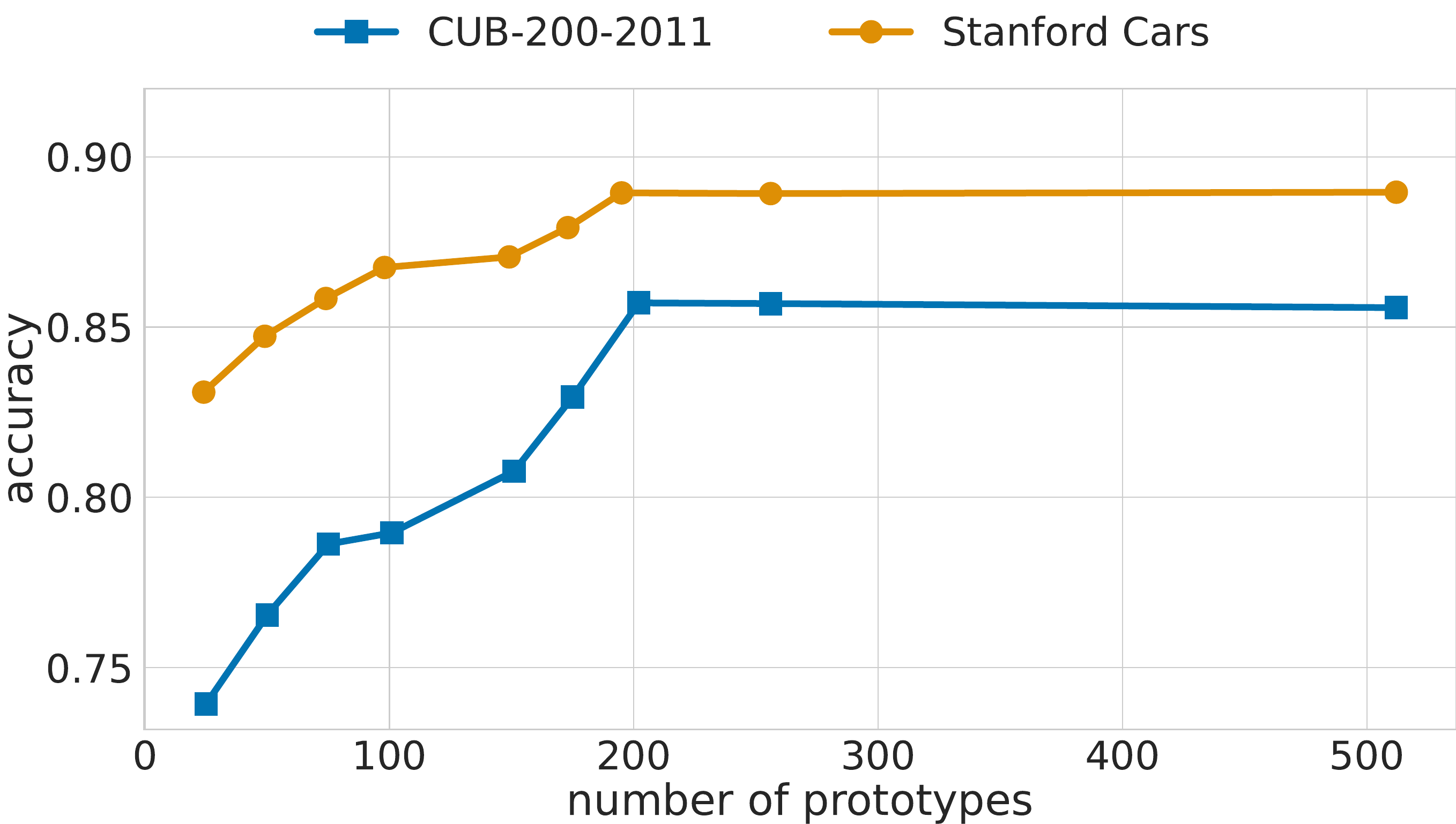}
        \caption{Accuracy depending on the number of prototypes. One can observe that the model reaches a plateau for around 200 prototypical parts, and there is no gain in further increase of prototype number.}
        \label{fig:proto_number}
     \end{subfigure}
     \hfill
     \begin{subfigure}[t]{0.49\textwidth}
         \centering
        \includegraphics[width=\textwidth]{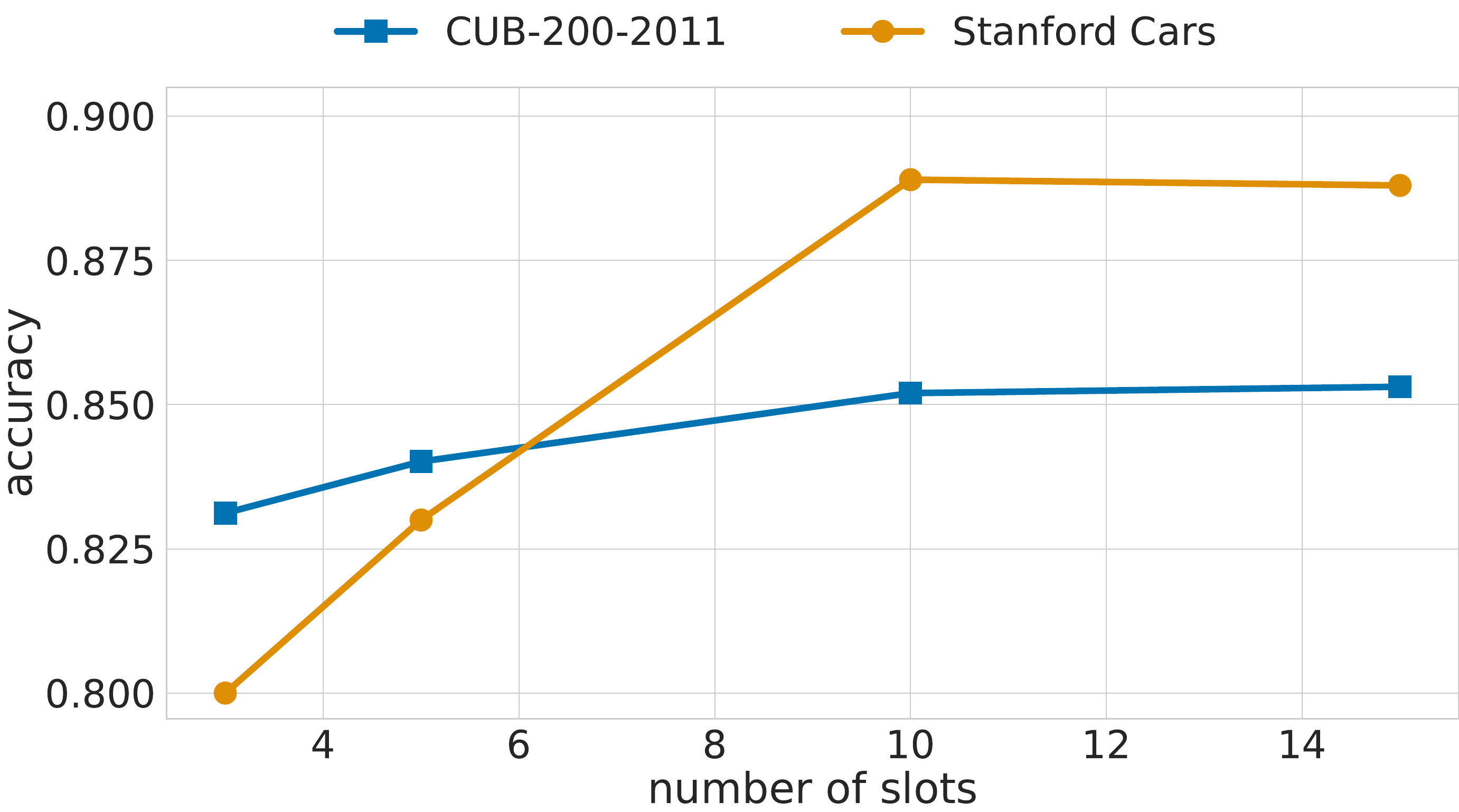}
        \caption{Accuracy depending on the number of slots. One can observe that the model reaches a plateau for around $10$ slots per class.}
    \label{fig.slot_number}
     \end{subfigure}
        \caption{ProtoPool accuracy with ResNet50 backbone depending on the number of prototypes and slots for CUB-200-2011 (blue square) and  Stanford Cars (orange circle) datasets.}
        \label{fig:protopool_number}
\end{figure}

\paragraph{Before and after prototype projection}
Since ProtoPool has much fewer prototypical parts than other models based on a positive reasoning process, applying projection could result in insignificant prototypes and reduced model performance. Therefore, we decided to test model accuracy before and after the projection (see \Cref{tab:push}), and we concluded that differences are negligible.

\paragraph{Number of prototypes and slots vs accuracy}
Finally, in \Cref{fig:protopool_number} we investigate how the number of prototypical parts or slots influences accuracy for the CUB-200-2011 and Stanford Cars datasets. We observe that up to around 200 prototypical parts, the accuracy increases and reaches the plateau. Therefore, we conclude that the amount of prototypes optimal for ProtoTree is also optimal for ProtoPool. Similarly, in the case of slots, ProtoPool accuracy increases till the $10$ slots and then reaches the plateau. 


\section{Conclusions}

We presented ProtoPool, a self-explainable method that incorporates the paradigm of prototypical parts to explain its predictions. This model shares the prototypes between classes without pruning operations, reducing their number up to ten times. Moreover, it is fully differentiable. To efficiently assign the prototypes to classes, we apply the Gumbel-Softmax trick together with orthogonalization loss. Additionally, we introduced focal similarity that focuses on salient features. As a result, we increased the interpretability while maintaining high accuracy, as we showed through theoretical analysis, multiple experiments, and user study.

\section*{Acknowledgments}
The research of J. Tabor, K. Lewandowska and D. Rymarczyk were carried out within the research project ``Bio-inspired artificial neural network'' (grant no. POIR.04.04.00-00-14DE/18-00) within the Team-Net program of the Foundation for Polish Science co-financed by the European Union under the European Regional Development Fund. The work of Ł. Struski and B. Zieliński were supported by the National Centre of Science (Poland) Grant No. 2020/39/D/ST6/01332, and 2021/41/B/ST6/01370, respectively. 

{\small
\bibliographystyle{splncs04}
\bibliography{egbib}
}

\section*{Supplementary Materials}
\section{Details on experimental setup}

We use two datasets, CUB-200-2011~\cite{wah2011caltech} consisted of 200 species of birds and Stanford Cars~\cite{krause20133d} with 196 car models. For both datasets, images are augmented offline using parameters from~\Cref{tab:Augmentation}, and the process of data preparation is the same as in~\cite{chen2019looks}\footnote{see \textit{Instructions for preparing the data} at \url{https://github.com/cfchen-duke/ProtoPNet}}.

\begin{table}[]
    \centering
    \caption{Augmentation policy.}
    \small
    \begin{tabular}{lcc}
    \toprule
         Augmentation & Value & Probability \\
         \midrule
         Rotation & $[-15^{\circ},15^{\circ}]$ & $1.0$ \\
         Flip & Vertical &  $0.5$ \\
         Flip & Horizontal &  $0.5$ \\
         Skew & $<45^{\circ}$ &  $0.5$ \\
         Shear &  $[-10^{\circ},10^{\circ}]$ &  $1.0$ \\
         Mix-up~\cite{thulasidasan2019on} & $50\%:50\%$  & $1.0$ \\
         \bottomrule
    \end{tabular}
    
    \label{tab:Augmentation}
\end{table}

Our model consists of the convolutional part $f$ that is a convolutional block from ResNet or DenseNet followed by $1\times1$ convolutional layer required to transform the latent space depth to $128$ for Stanford Cars and $256$ for CUB-200-2011.
We perform a warmup training where the weights of $f$ are frozen for $10$ epochs, and then we train the model until it converges with $12$ epochs early stopping. After convergence, we perform prototype projection and fine-tune the last layer. We use the learning schema presented in~\Cref{tab:learning_scheme}.

\begin{table}[]
    \centering
    \caption{Learning schema for the ProtoPool model.}
    \fontsize{8}{10}\selectfont
    \begin{tabular}{l@{\;}ccccc}
    \toprule
         Phase & Model layers & Learning rate & Scheduler & Weight decay & Duration \\
         \midrule
         \multirow{2}{*}{Warm-up} & add-on $1\!\!\times\!\!1$ convolution & $1.5\cdot 10^{-3}$ & \multirow{2}{*}{None} & \multirow{2}{*}{None} & \multirow{2}{*}{$10$ epochs}\\
         & prototypical pool & $1.5\cdot 10^{-3}$ \\
         \midrule
         \multirow{3}{*}{Joint} & convolutions $f$ & $5\cdot 10^{-5}$ & \multirow{3}{*}{\shortstack{by half every \\ $5$ epochs}} & \multirow{3}{*}{$10^{-3}$} & \multirow{3}{*}{\shortstack{$12$ epoch \\ early stopping}} \\
         & add-on $1\!\!\times\!\!1$ convolution & $1.5\cdot 10^{-3}$ & & & \\
         & prototypical pool & $1.5\cdot 10^{-3}$ & &  & \\
         \midrule
         After & \multirow{2}{*}{last layer} & \multirow{2}{*}{$10^{-4}$} & \multirow{2}{*}{None} & \multirow{2}{*}{None} & \multirow{2}{*}{$15$ epochs} \\
         projection & \\
         \bottomrule
    \end{tabular}
    
    \label{tab:learning_scheme}
\end{table}

Additionally, we employ Adam optimizer~\cite{kingma2015adam} with parameters $\beta_1=0.9$ and $\beta_2=0.999$. We set the batch size to $80$ and use input images of resolution $224\times224\times3$. Moreover, we use prototypical parts of size $1\times 1 \times 128$ and $1\times 1 \times 256$ for Stanford Cars and CUB-200-2011 respectively.
The weights between the class logit and its slots are initialized to $1$, while the remaining weights of the last layer are set to $0$. All other parameters of the network are initialized with Xavier's normal initializer.

We utilize the Gumbel-Softmax trick to unambiguously assign prototypes to class slots. However, in contrast to the classic variant of this parametrization trick, we reduce the influence of the noise in subsequent iterations. For this purpose, we use
\(
y^i = \tfrac{\exp\left(q^i / \tau + \eta_i\right)}{\sum^M_{m=1}\exp\left(q^m / \tau + \eta_m\right)}
\)
instead of
\(
y^i = \tfrac{\exp\left((q^i + \eta_i) / \tau\right)}{\sum^M_{m=1}\exp\left((q^m + \eta_m) / \tau\right)}
\)
in Gumbel-Softmax distribution
\[
\addtolength\abovedisplayskip{-0.2\baselineskip}%
\addtolength\belowdisplayskip{-0.2\baselineskip}%
\mathrm{Gumbel\textnormal{-}softmax}(q, \tau) = (y^1, \ldots, y^M)\in\R^M,
\]
where $\tau\in (0, \infty)$ and $q\in \mathbb{R}^M$.
Moreover, we start the Gumbel-Softmax distribution with $\tau=1$, decreasing it to $0.001$ for $30$ epochs. As a decrease function, we use
\[
\tau(epoch) = 
\begin{cases} 
1 / \sqrt{\alpha\cdot\text{epoch}} & \text{if } \text{epoch} < $30$ \\
0.001 & \text{otherwise} 
\end{cases},
\]
where $\alpha = 3.4\cdot 10^4$. 
We use the following weighting schema for loss function: $\mathcal{L}_{\small entropy} = 1.0$, $\mathcal{L}_{\small clst} = 0.8$, $\mathcal{L}_{\small sep} = -0.08$, $\mathcal{L}_{\small orth} = 1.0$, and $\mathcal{L}_{\small l_1} = 10^{-4}$. Finally, we normalize $\mathcal{L}_{\small orth}$, dividing it by the number of classes multiplied by the number of slots per class.

\section{Generating names of prototypical parts}

To name the prototype for CUB-200-2011, we used the attributes of images collected with Amazon Mechanical Turk that are provided together with the dataset. Firstly, for a given image, we filter out attributes assigned by less than 20\% users. Then, for each class, we remove attributes present at less than 20\% of testing images assigned to this class. Later, we use five nearest patches of a given prototype to determine if they are consistent and point to the same part of the bird. Eventually, we choose the attributes accurately describing the nearest patches for a given prototype (see Figure~1 from the main paper).

\section{Results for other backbone networks}

\begin{figure}[t]
\centering
    \centering
    \includegraphics[width=0.8\textwidth]{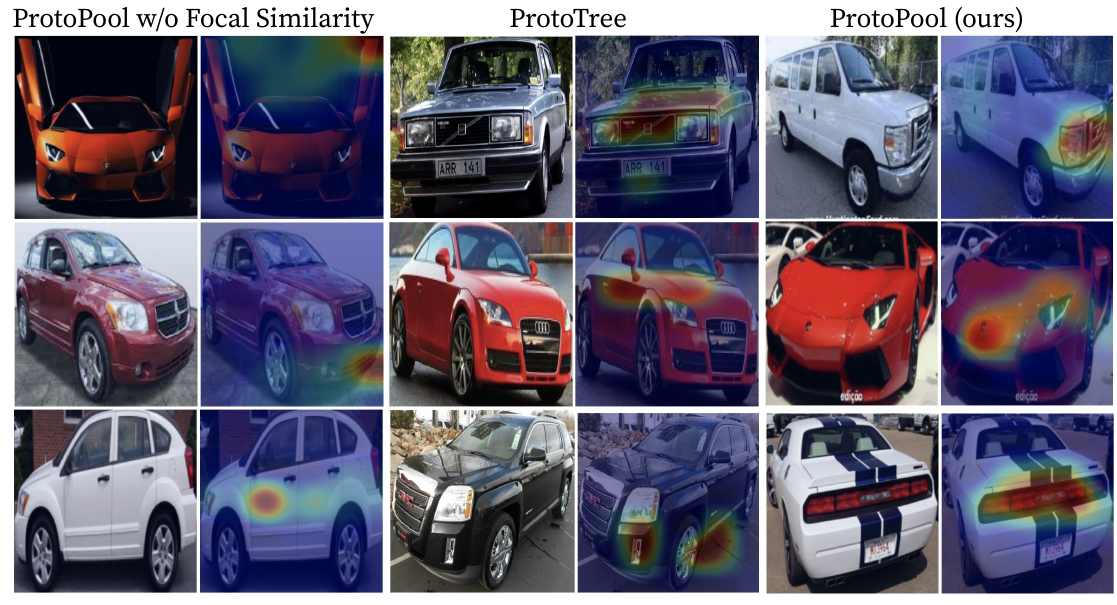}
    \caption{Sample prototype activation maps obtained with models using various similarity functions. One can observe that ProtoPool with the focal similarity is the only one that focuses on car features like a spoiler, reflector, and \textit{Ferrari} logo when the other similarities focus on image borders or much larger regions. Since similarity function is a design choice of the model, the images from which the prototypical parts derive are different.}
    \label{fig:diff_sim}
\end{figure}

In \Cref{tab:sup_acc}, we present the results for ProtoPool with DenseNet as a backbone network. Moreover, in \Cref{fig:diff_sim}, we present examples of prototypes derived from three different methods: ProtoPool, ProtoTree, and ProtoPool without focal similarity.

\begin{table}[]
  \caption{Comparison of ProtoPool with other methods based on prototypical parts trained on the CUB-200-2011 and Stanford Cars datasets, considering a various number of prototypes and types of convolutional layers $f$. ProtoPool achieves competitive results, even for models with ten times more prototypes in the case of both datasets. Please note that the results are first sorted by backbone network and then by the number of prototypes.}
  \centering
  \small
  \begin{tabular}{@{}c@{\;\;}l@{\;}c@{\;\;}cc@{}}
    \toprule
    Data & \multicolumn{1}{c}{Model} & Architecture & Proto. \# & Acc [\%] \\
    \midrule
    \multirow{8}{*}{\rotatebox{90}{\textbf{CUB-200-2011}}} & ProtoPool (ours) & \multirow{4}{*}{DenseNet121}  & 202 & $73.6 \pm 0.4$ \\
    & ProtoPShare~\cite{rymarczyk2021protopshare} &  & $600$ & $74.7$ \\
    & ProtoPNet~\cite{chen2019looks} & & $1476$ & $79.2$\\
    & TesNet~\cite{wang2021interpretable} &  & $2000$ & $\mathbf{84.8 \pm 0.2}$\\
    \cmidrule{2-5}
    & ProtoPool (ours) & \multirow{4}{*}{DenseNet161}  & 202 & $80.3 \pm 0.3$ \\
    & ProtoPShare~\cite{rymarczyk2021protopshare} &  & $600$ & $76.5$ \\
    & ProtoPNet~\cite{chen2019looks} & & $1527$ & $79.9$\\
    & TesNet~\cite{wang2021interpretable} &  & $2000$ & $84.6 \pm 0.3$\\
    \cmidrule{1-5}
    \multirow{4}{*}{\rotatebox{90}{\textbf{Cars}}}& ProtoPool (ours) & \multirow{4}{*}{DenseNet121} & $202$ & $86.4 \pm 0.1$ \\
    & ProtoPShare~\cite{rymarczyk2021protopshare} &  & $980$ & $84.8$ \\
    & ProtoPNet~\cite{chen2019looks} &  & $2000$ & $86.8\pm0.1$\\
    & TesNet~\cite{wang2021interpretable} &  & $2000$ & $\mathbf{92.0 \pm 0.3}$\\
   \bottomrule
  \end{tabular}
  
  \label{tab:sup_acc}
\end{table}

\begin{table}[]
  \captionsetup{skip=2pt}
  \centering
  \caption{Comparison of prototypical methods for fine-grained image classification. One can observe that ProtoPool uses only 10\% of ProtoPNet's prototypes, remaining interpretable due to the positive reasoning process. Additionally, ProtoPool, similarly to the ProtoPShare, detects inter-class similarity and shares the prototypes. But, it is trained end-to-end thanks to the differentiable prototypes assignment. On the other hand, ProtoTree is the only model presenting the explanation in a hierarchical way, which requires more time to comprehend according to human cognitive system theory~\cite{fiske1991social}.}
  \label{tab:full_interpretability}
  \fontsize{8}{10}\selectfont
  \begin{tabular}{@{}l@{\quad}c@{\quad}c@{\;\;}c@{\quad}c@{\;\;}c@{\;\;}c@{}}
    \toprule
    \multirow{2}{*}{Model} &  \multirow{2}{*}{Proto. \#} & Diff. proto. & Information & \multirow{2}{*}{\shortstack{Reasoning\\ type}} & \multirow{2}{*}{\shortstack{Proto.\\ sharing}} & \multirow{2}{*}{\shortstack{Class\\ similarity}}  \\
    & & assignment & processing & &\\
    \midrule
    ProtoPNet & 100\% & no &  simultaneous & positive & none & none \\
    TesNet & 100\% & no & simultaneous & positive & none & none \\
    ProtoPShare & [20\%;50\%] & no & simultaneous & positive & direct & direct \\
    ProtoTree & 10\% & no & successive & positive/negative & indirect & indirect \\
    ProtoPool & 10\% & yes & simultaneous & positive & direct & direct\\
    \bottomrule
  \end{tabular}
\end{table}

\section{Comparison of the prototypical models}

\begin{figure}[tp]
    \centering
    \includegraphics[width=0.48\textwidth]{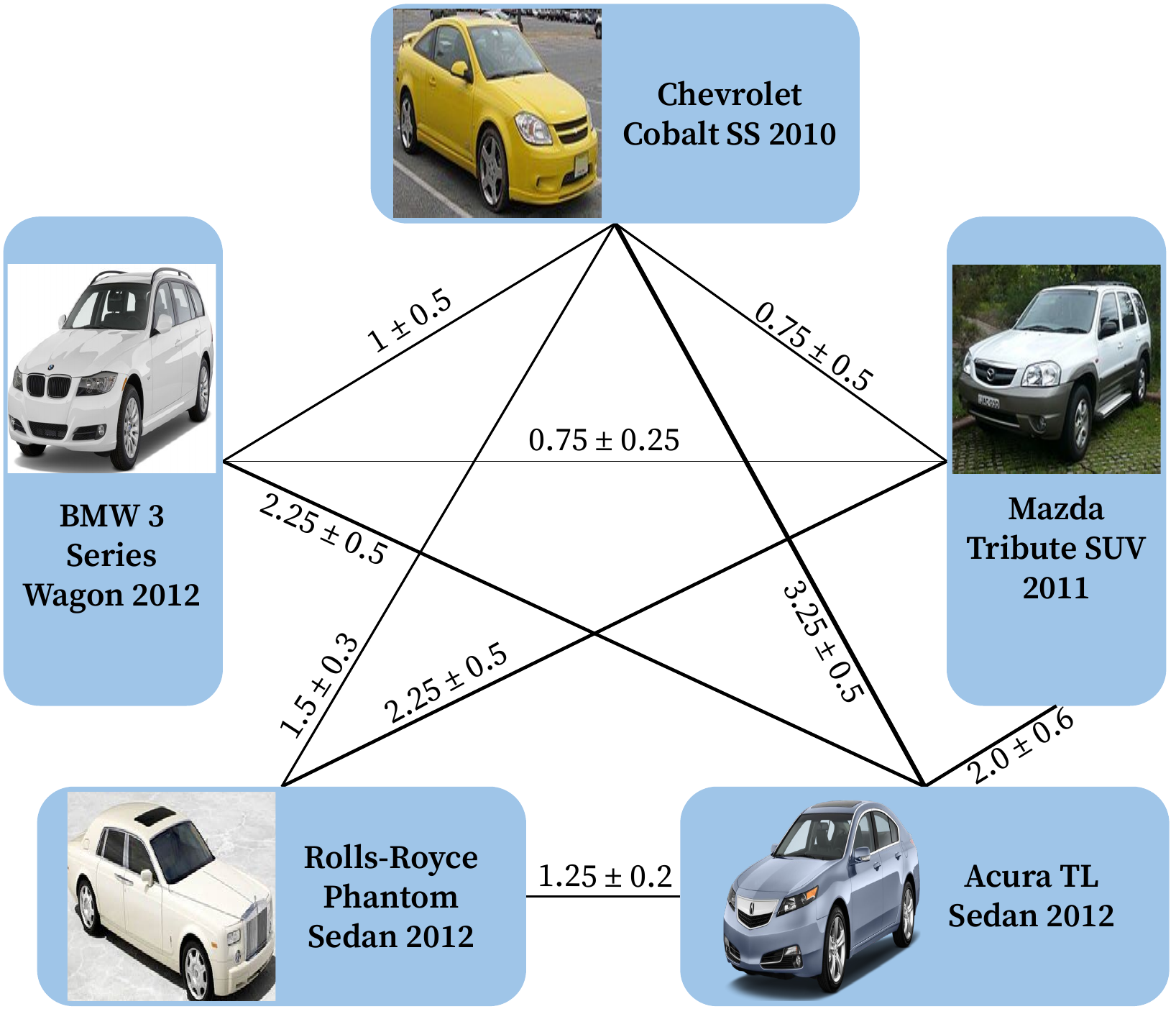}
    \caption{Sample graph obtained from  ProtoPool for five classes from the Stanford Cars dataset. Each class is represented by a single image, and the edges correspond to the mean number of prototypes shared between classes over five repetitions. One can observe that ProtoPool discovered similarities between SUVs and Sedans, while Rolls-Royce and BMW have nothing in common. Moreover, the graphs are consistent between runs and discover similar relations between classes.}
    \label{fig:graph_sharing}
\end{figure}

In \Cref{tab:full_interpretability}, we present the extended version of prototypical-based model comparison. One can observe that ProtoPool is the only one that has a differentiable prototypical parts assignment. Additionally, it processes the data simultaneously, which is faster and easier to comprehend by humans~\cite{fiske1991social} than sequential processing obtained from the ProtoTree~\cite{nauta2020looks}. Lastly, ProtoPool, as ProtoPShare, directly provides class similarity that is visualized in the~\Cref{fig:graph_sharing}.

\section{Details on ablation study}

\begin{figure*}[]
    \captionsetup{skip=2pt}
    \centering
    \begin{subfigure}{0.49\textwidth}
        \includegraphics[width=\textwidth]{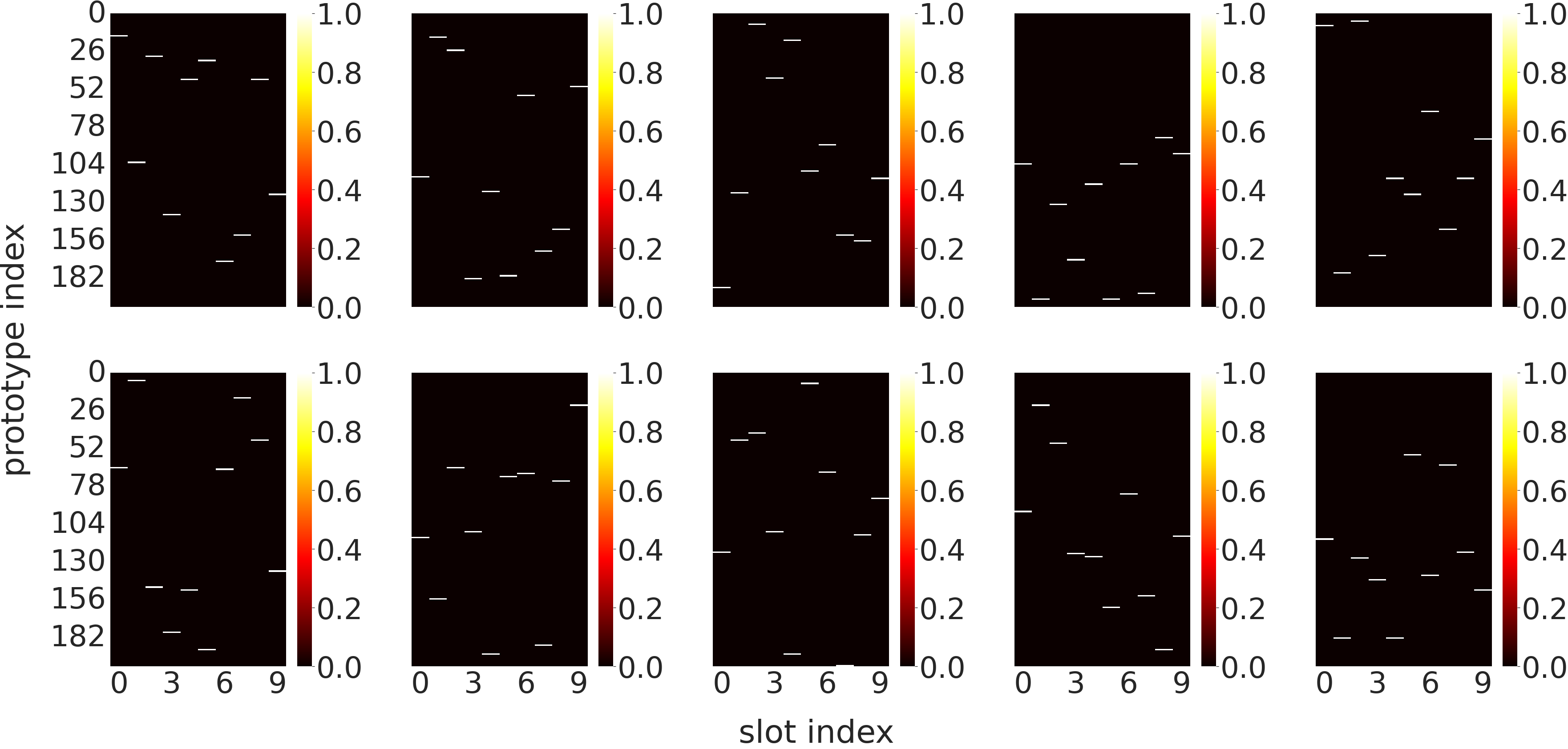}
        \caption{ProtoPool}
        \label{fig.all_birds}
    \end{subfigure}
    \hfill
    \begin{subfigure}{0.49\textwidth}
        \includegraphics[width=\textwidth]{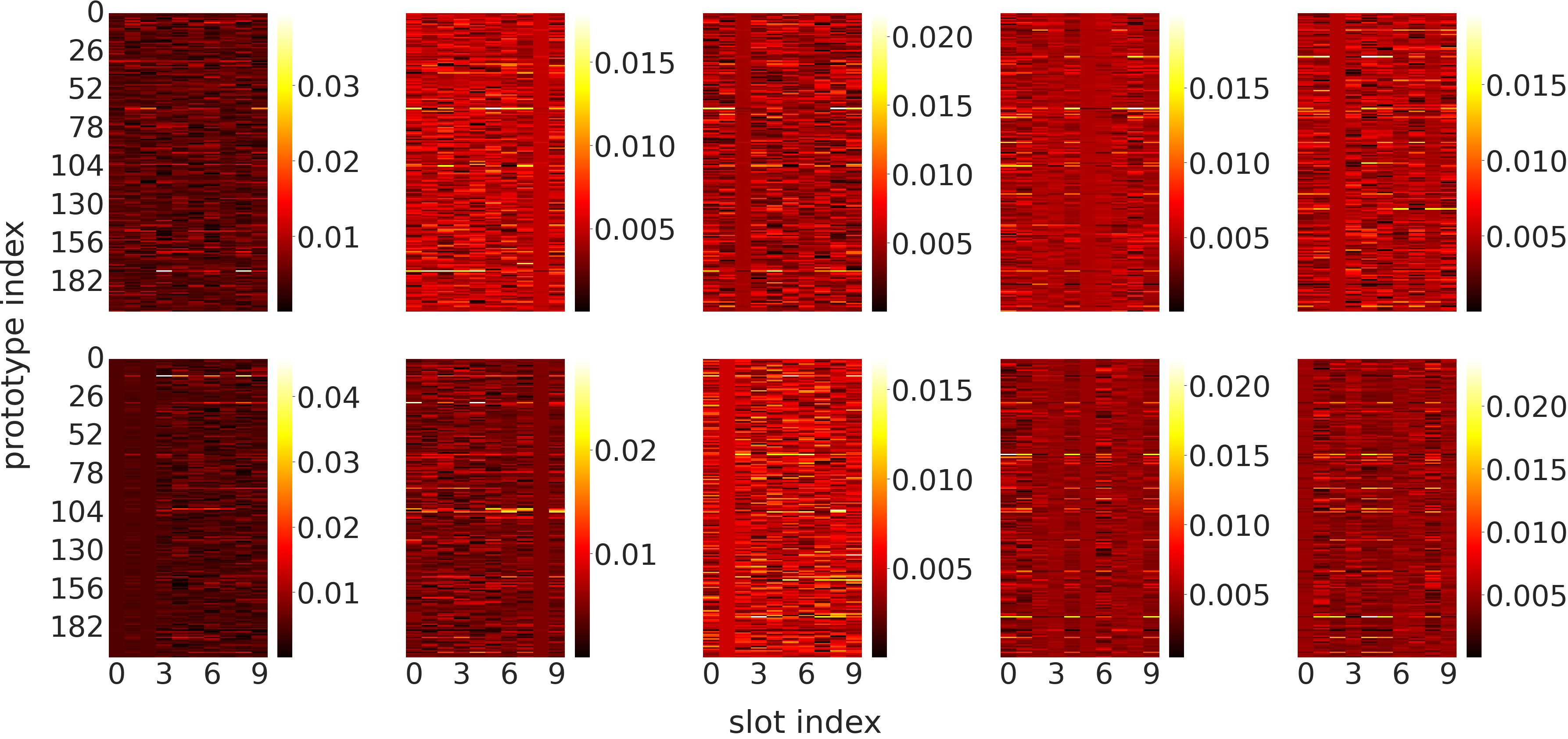}
        \caption{w/o Gumbel-Softmax trick}
        \label{fig.no_gumb}
    \end{subfigure}
    \\
    \begin{subfigure}{0.49\textwidth}
        \includegraphics[width=\textwidth]{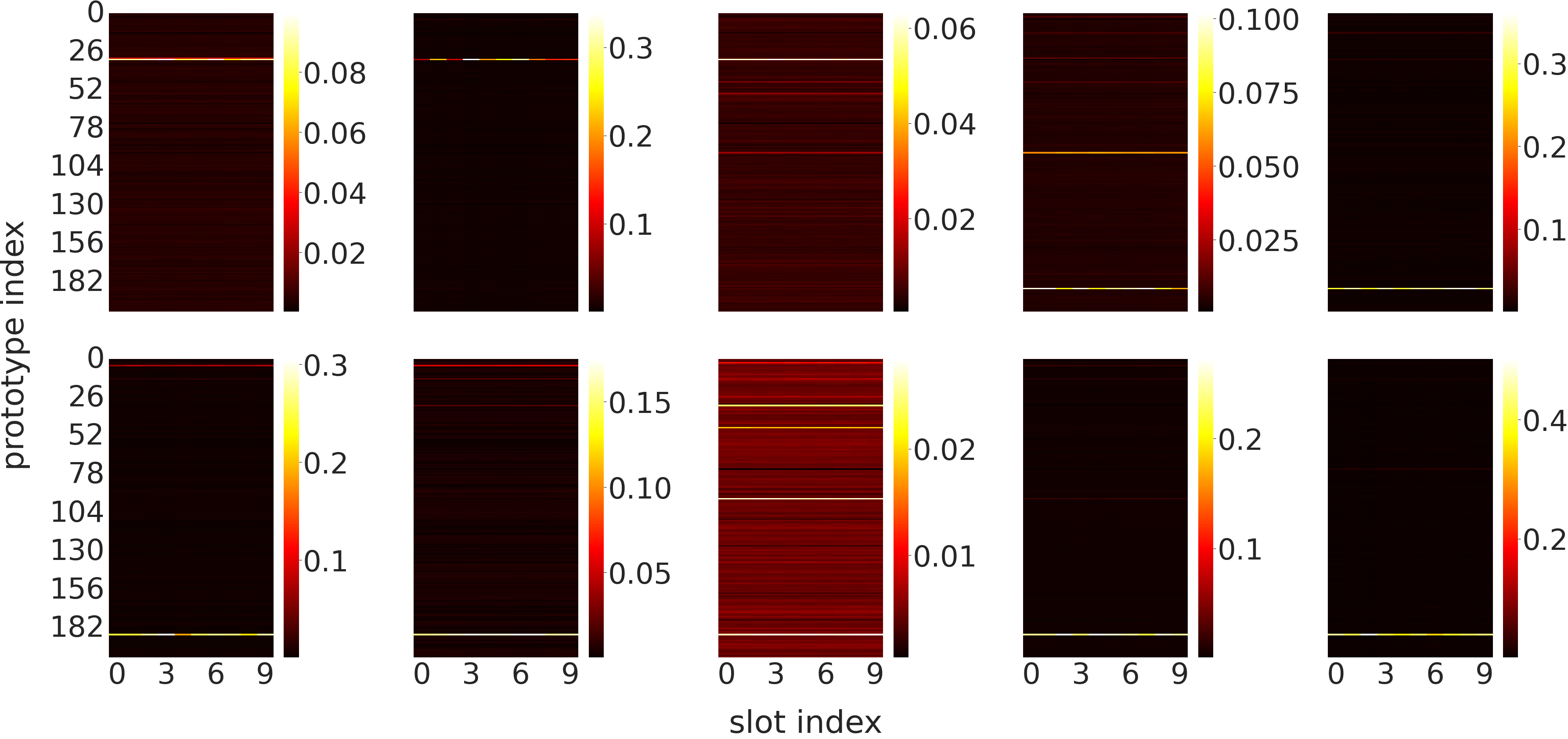}
        \caption{w/o $\mathcal{L}_{\small orth}$}
        \label{fig.no_orth}
    \end{subfigure}
    \hfill
    \begin{subfigure}{0.49\textwidth}
        \includegraphics[width=\textwidth]{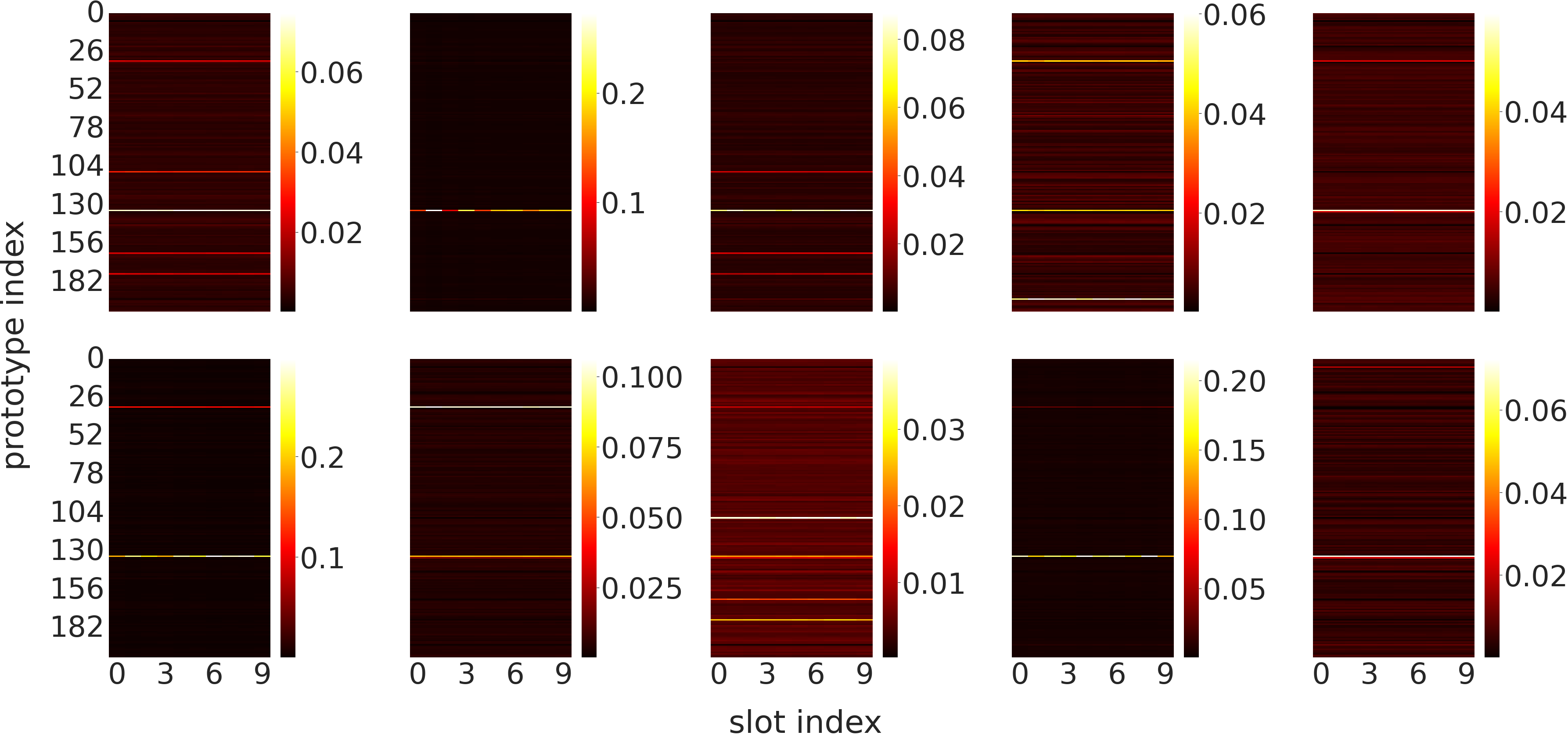}
        \caption{w/o $\mathcal{L}_{\small orth}$ and Gumbel-Softmax trick}
        \label{fig.none}
    \end{subfigure}
    \begin{subfigure}{0.49\textwidth}
        \includegraphics[width=\textwidth]{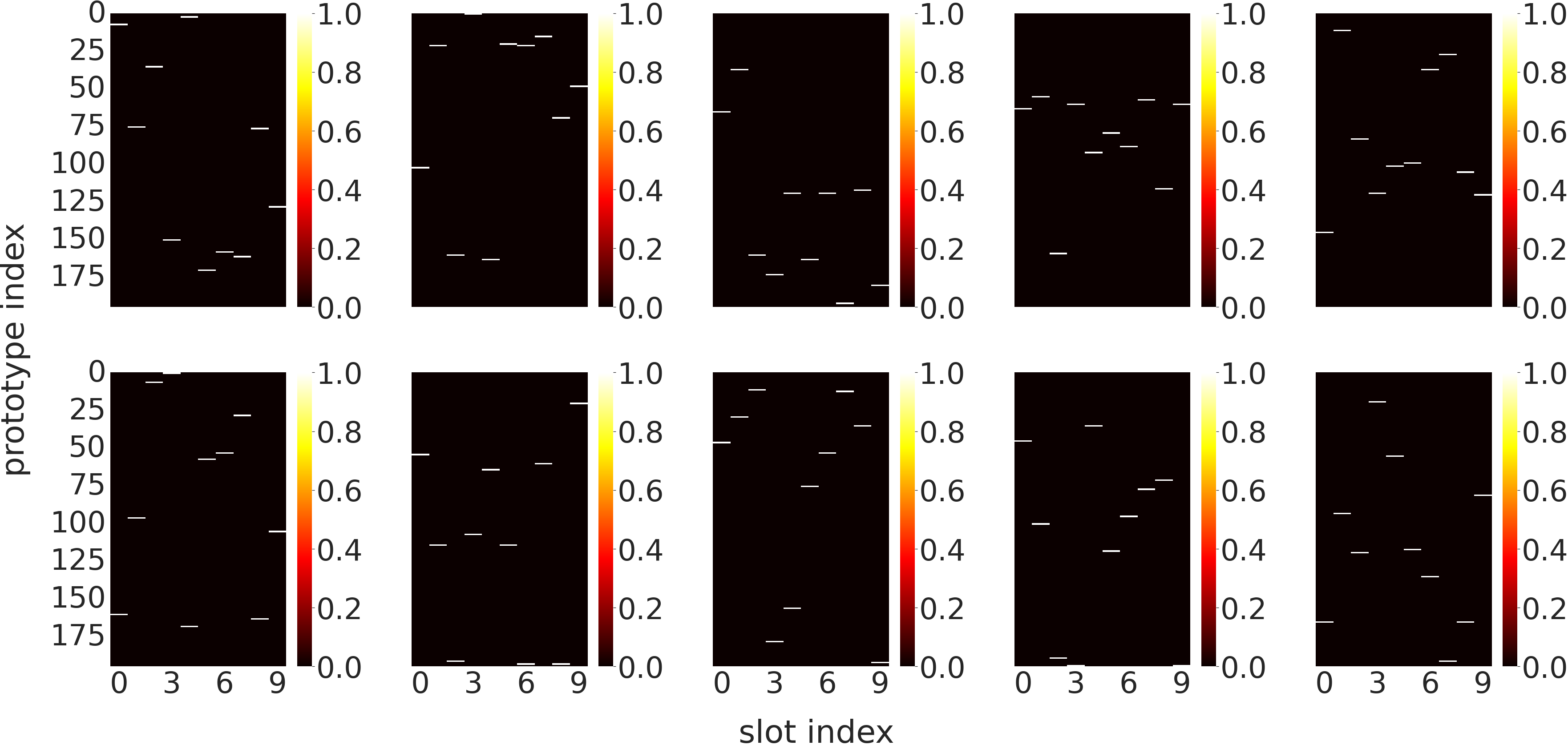}
        \caption{ProtoPool}
        \label{fig.cd_diagram}
    \end{subfigure}
    \hfill
    \begin{subfigure}{0.49\textwidth}
        \includegraphics[width=\textwidth]{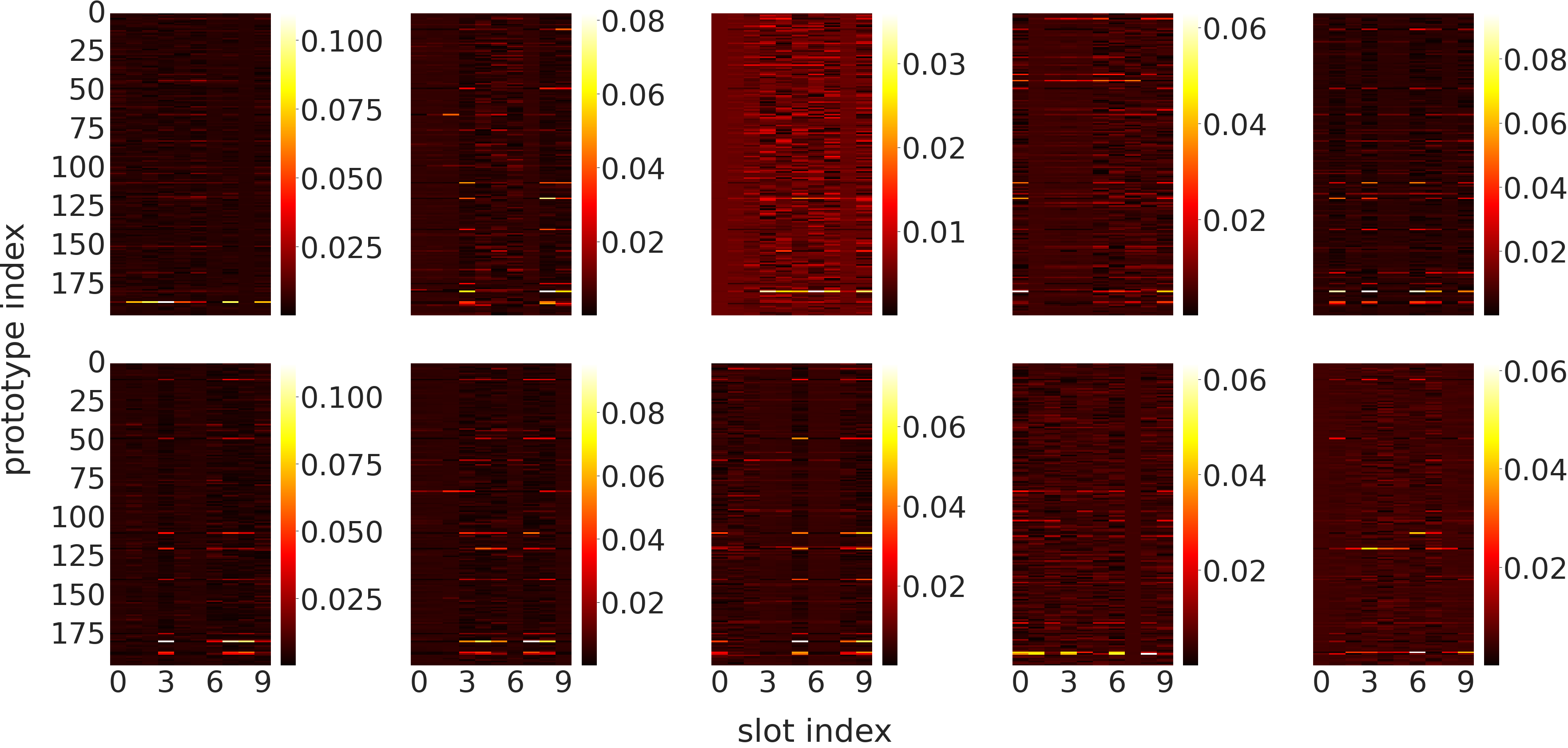}
        \caption{w/o Gumbel-Softmax trick}
        \label{fig.user_study_results}
    \end{subfigure}
    \\
    \begin{subfigure}{0.49\textwidth}
        \includegraphics[width=\textwidth]{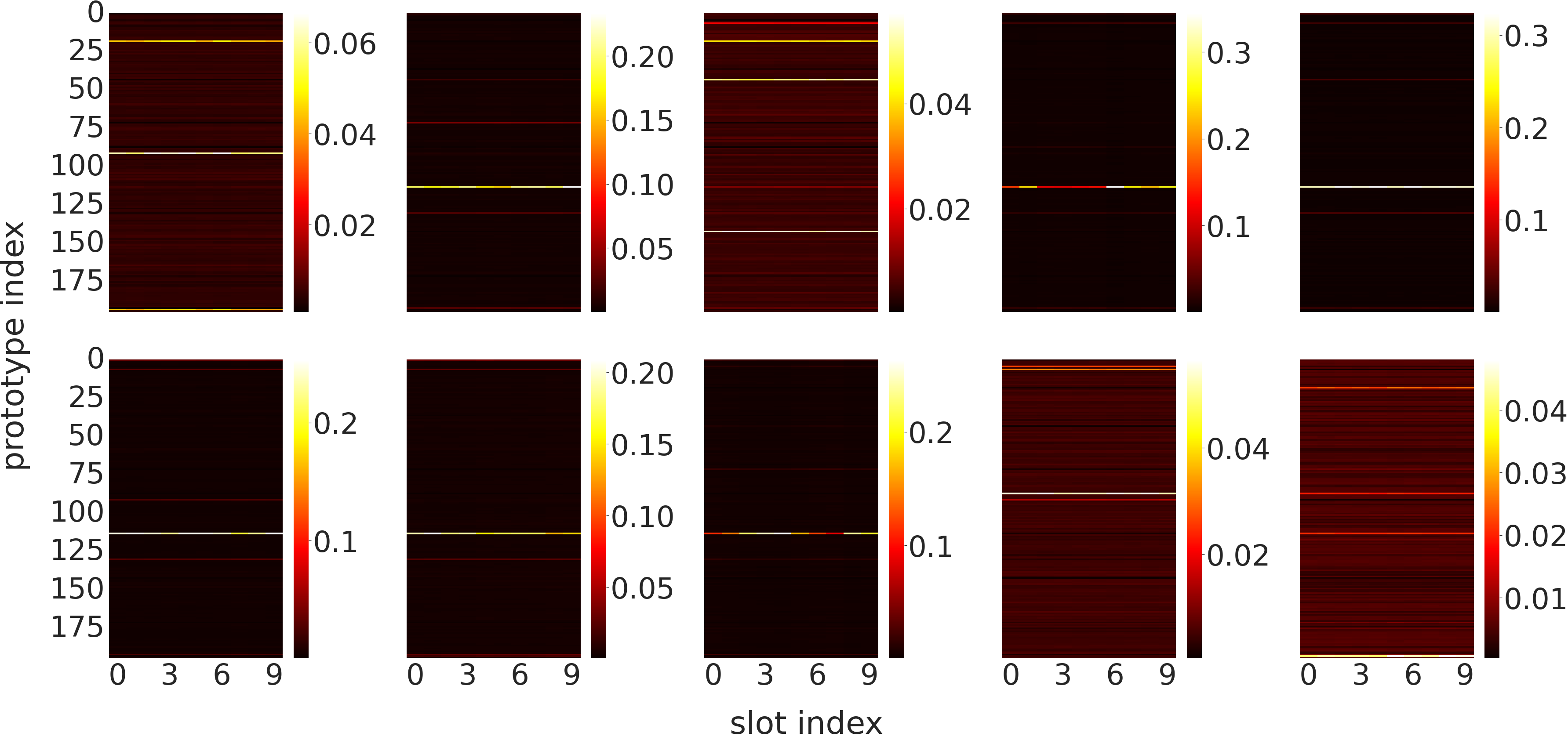}
        \caption{w/o $\mathcal{L}_{\small orth}$}
        \label{fig.user_study_results}
    \end{subfigure}
    \hfill
    \begin{subfigure}{0.49\textwidth}
        \includegraphics[width=\textwidth]{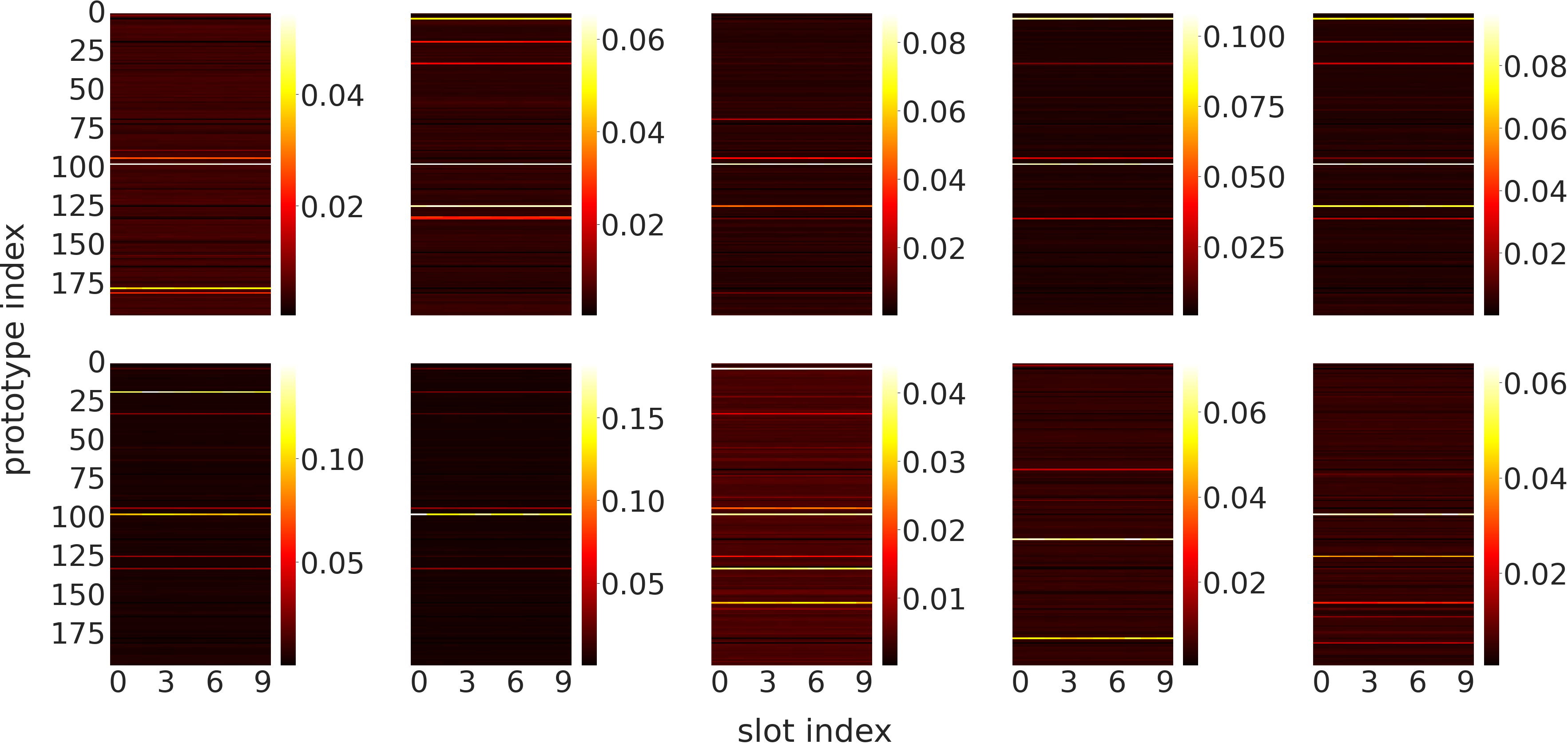}
        \caption{w/o $\mathcal{L}_{\small orth}$ and Gumbel-Softmax trick}
        \label{fig.user_study_results}
    \end{subfigure}
    
    \caption{The influence of novel architectural changes on prototypes to slots assignments for ten randomly chosen classes of the CUB-200-2011 (a-d) and Stanford Cars (e-h) datasets. Each class has ten slots (corresponding to columns) to which a prototype (corresponding to rows) can be assigned. As observed, the binarization of the assignment (hard assignment of a prototype) is obtained only for a mix of Gumbel-Softmax and $L_{orth}$. Moreover, if one of those factors is missing, the assignment matrix is random or aims to assign the same prototype to all slots. Note that the scale of heatmap colors differs between examples for clarity.}
    \label{fig.heatmap_birds}
\end{figure*}

\begin{figure*}[]
    \captionsetup{skip=2pt}
    \centering
    \begin{subfigure}{0.47\textwidth}
        \includegraphics[width=\textwidth]{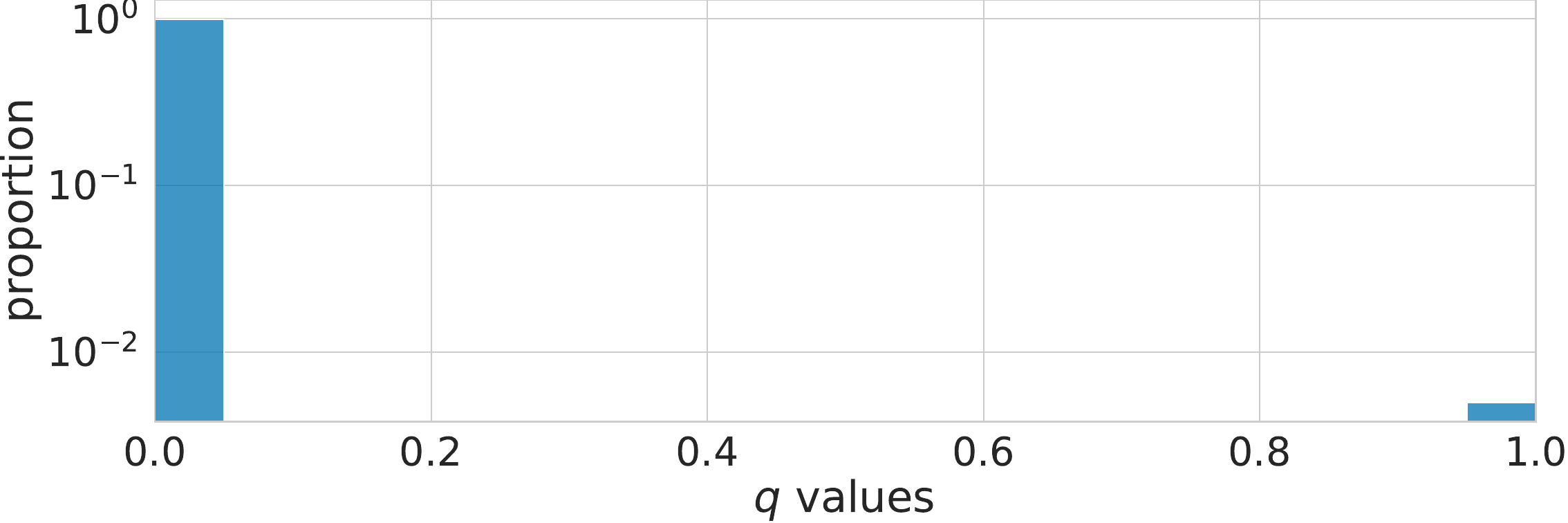}
        \caption{ProtoPool}
        \label{fig.cd_diagram}
    \end{subfigure}
    \hfill
    \begin{subfigure}{0.47\textwidth}
        \includegraphics[width=\textwidth]{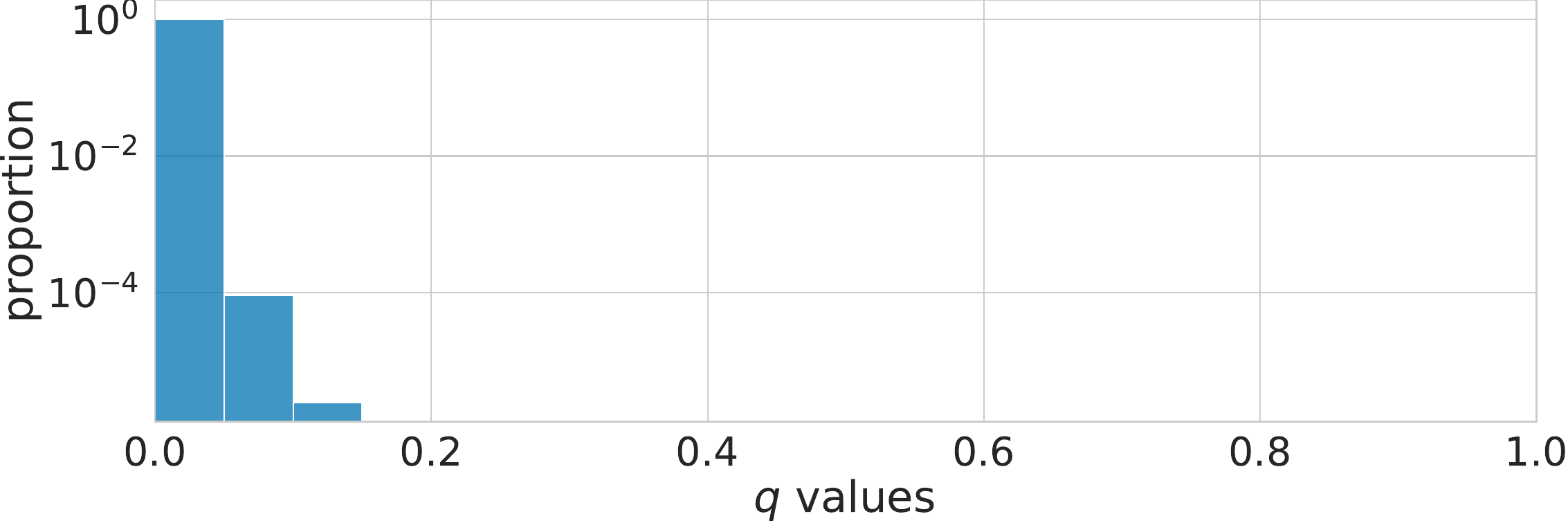}
        \caption{w/o Gumbel-Softmax trick}
        \label{fig.user_study_results}
    \end{subfigure}
    \hfill
    \begin{subfigure}{0.47\textwidth}
        \includegraphics[width=\textwidth]{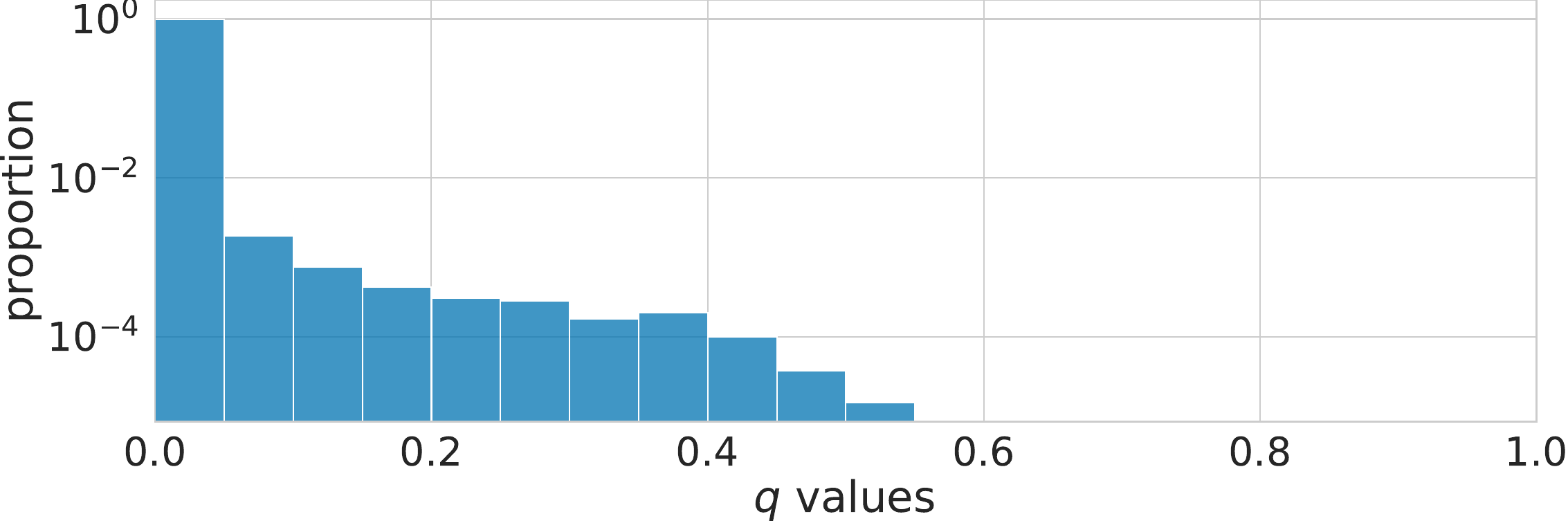}
        \caption{w/o $\mathcal{L}_{\small orth}$}
        \label{fig.user_study_results}
    \end{subfigure}
    \hfill
    \begin{subfigure}{0.47\textwidth}
        \includegraphics[width=\textwidth]{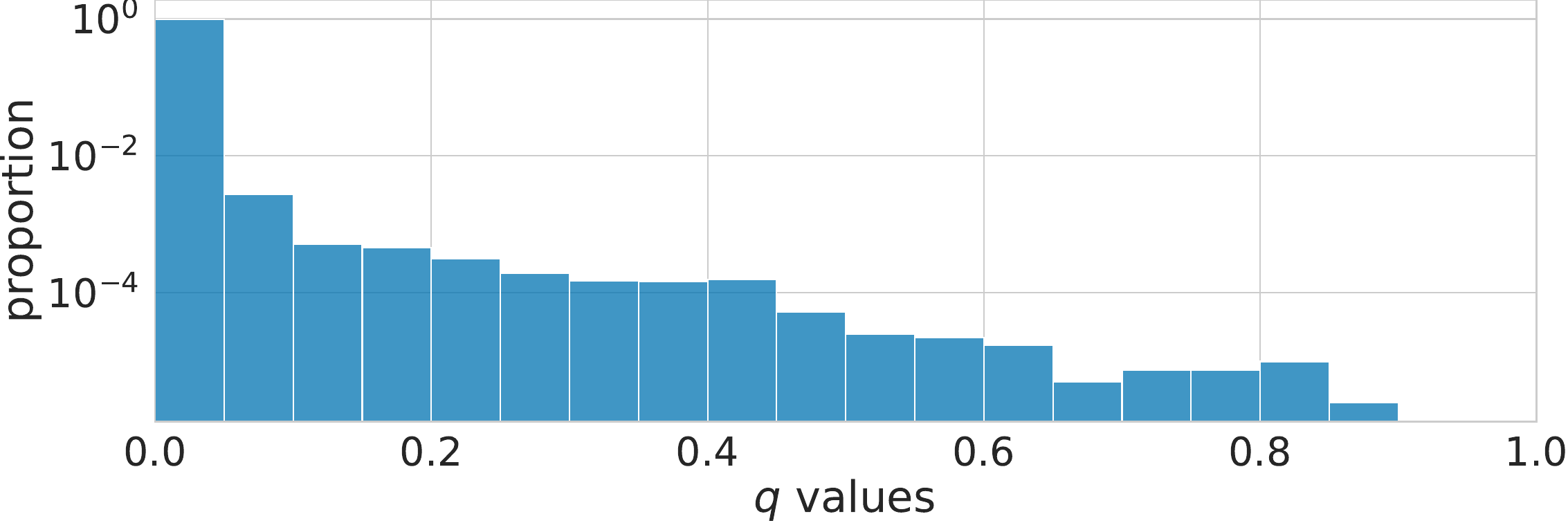}
        \caption{w/o $\mathcal{L}_{\small orth}$ and Gumbel-Softmax trick}
        \label{fig.user_study_results}
    \end{subfigure}
    \hfill 
    \begin{subfigure}{0.49\textwidth}
        \includegraphics[width=\textwidth]{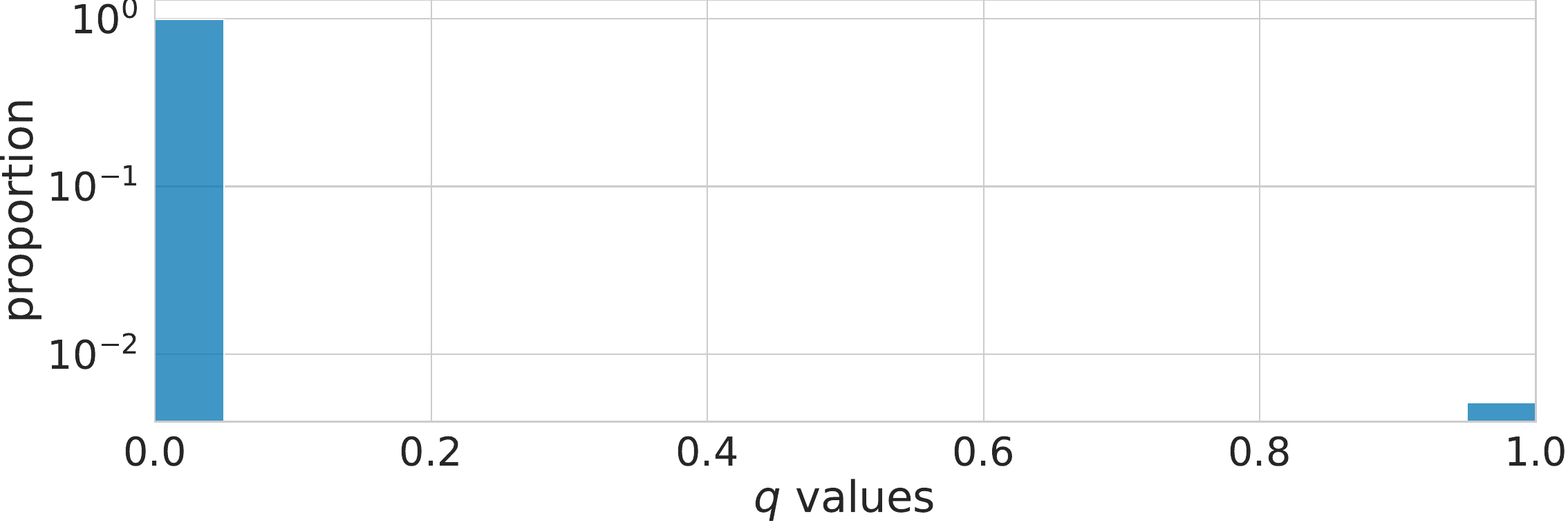}
        \caption{ProtoPool}
        \label{fig.cd_diagram}
    \end{subfigure}
    \hfill
    \begin{subfigure}{0.49\textwidth}
        \includegraphics[width=\textwidth]{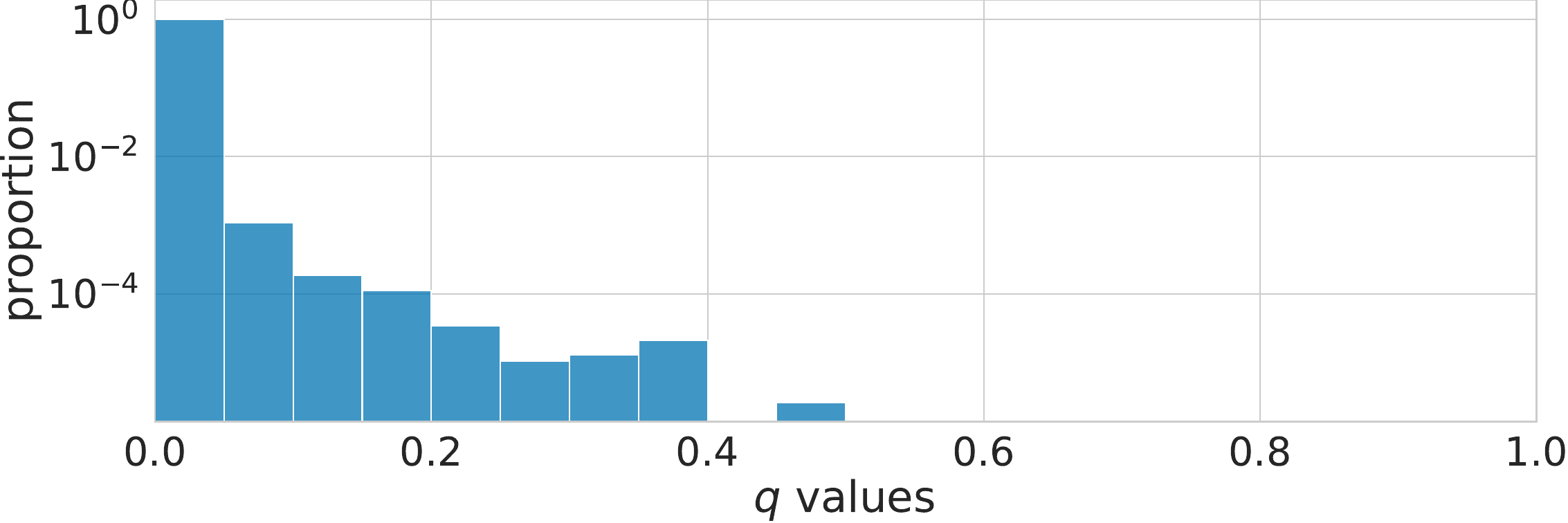}
        \caption{w/o Gumbel-Softmax trick}
        \label{fig.user_study_results}
    \end{subfigure}
    \hfill
    \begin{subfigure}{0.49\textwidth}
        \includegraphics[width=\textwidth]{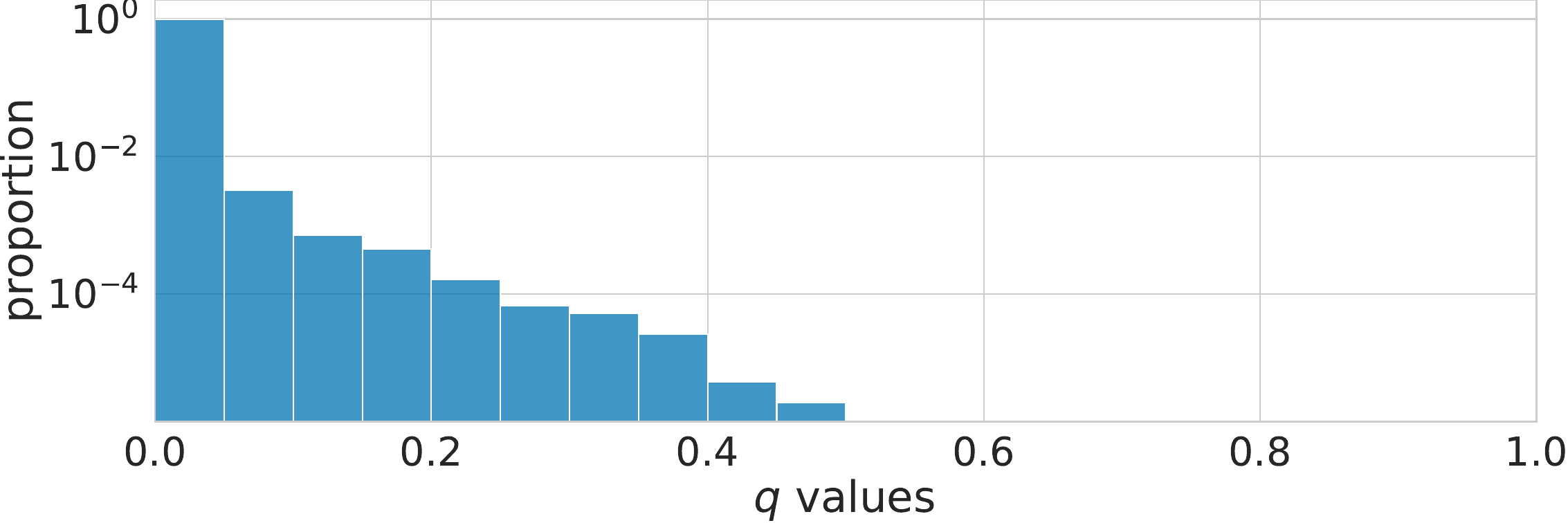}
        \caption{w/o $\mathcal{L}_{\small orth}$}
        \label{fig.user_study_results}
    \end{subfigure}
    \hfill
    \begin{subfigure}{0.49\textwidth}
        \includegraphics[width=\textwidth]{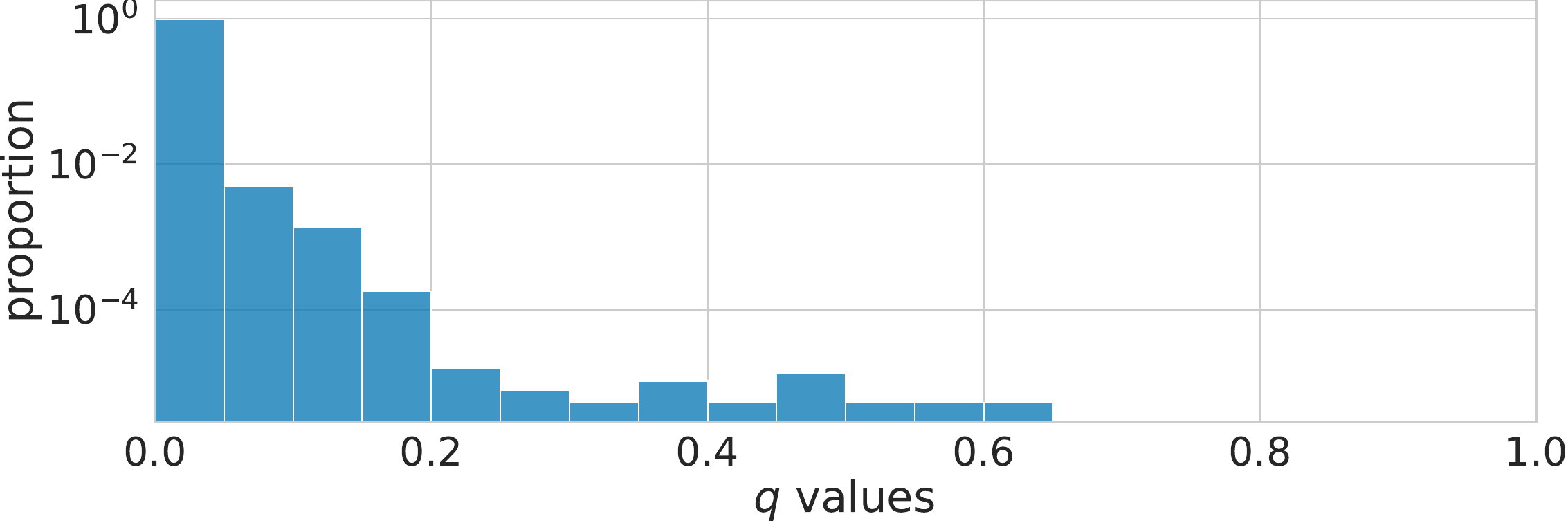}
        \caption{w/o $\mathcal{L}_{\small orth}$ and Gumbel-Softmax trick}
        \label{fig.user_study_results}
    \end{subfigure}
    \caption{The influence of novel architectural changes on the values of prototypes to slots assignments ($q$ distributions) for the CUB-200-2011 (a-d) and Stanford Cars (e-h) datasets. One can observe that only the ProtoPool model binarizes $q$ distributions. Note that the histograms are in logarithmic scale and normalized.}
    \label{fig.q_birds}
\end{figure*}

As an attachment to Table~4 from the main paper, in~\Cref{fig.heatmap_birds} we provide the matrices of prototype assignment. Moreover, in~\Cref{fig.q_birds}, we present the distribution of values from the prototype assignment matrix for corresponding datasets. As presented, only the ProtoPool model obtains bimodal distribution of 0 and 1, resulting in the binary matrix.

\section{Details on user study}

\begin{figure}
    \centering
    \includegraphics[width=0.9\textwidth]{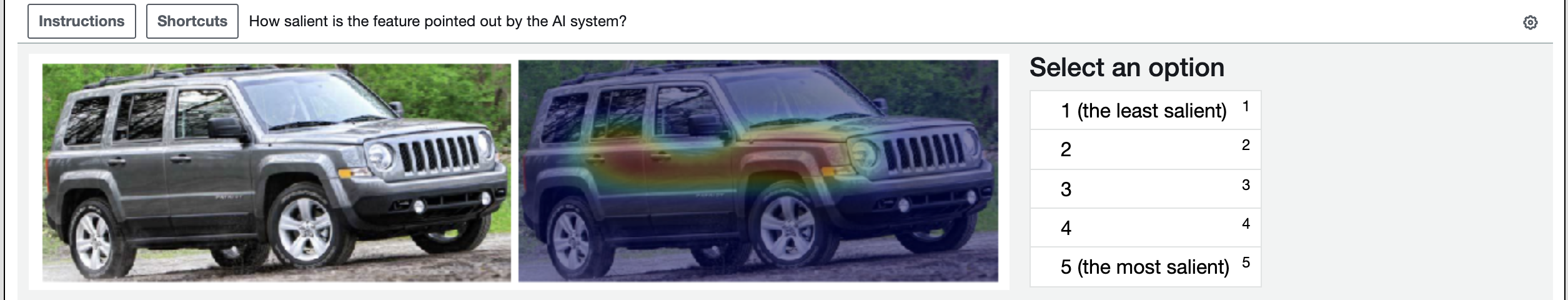}
    \caption{Sample question from the user study questionnaire.}
    \label{fig:question}
\end{figure}

To ensure a broad spectrum of the users, we ran the AMT batches at four different hours (8 AM, 2 PM, 8 PM, 12 PM CET) and required balanced sex in (53\% of women) and versatile age (from 20 to 60) of the users. Each user assessed examples of prototypical parts generated by ProtoPool, ProtoTree~\cite{nauta2021neural} and ProtoPool without focal similarity in a randomized order. The user did not know which image was generated by which model, and there was no difference in preprocessing those images between models.
Each person answered ten questions for each dataset and model combination, resulting in 60 responses per participant. These 60 images were randomly selected from the pool of 180 images (30 for each combination of dataset and model).
Each user had unlimited time for the answer. The task was to assign a score from 1 to 5 where 1 meant \textit{``Least salient''} and 5 meant \textit{``Most salient''}. A sample question is presented in~\Cref{fig:question} and the results are shown in ~\Cref{tab:user}.
One can observe that ProtoPool achieves the highest number of positive answers (4 and 5). Additionally, ProtoTree is better for Stanford Cars rather than CUB-200-2011, which can be correlated to the weaker intra-class similarity in the case of car models~\cite{nauta2021neural}. Overall, we conclude that the enrichment of the model with focal similarity substantially improves the model interpretability and better detects salient features.

\begin{table}[]
    \centering
    \caption{User study results show that only ProtoPool has the most positive votes (4 and 5) in both datasets. Additionally, ProtoTree is more interpretable for Stanford Cards than for CUB-200-2011.}
    \small
    \begin{tabular}{@{}lcccccc@{}}
    \toprule
         \multirow{2}{*}{Model} & \multirow{2}{*}{Dataset} & \multicolumn{5}{c}{Answers}\\

         & & 1 & 2 & 3 & 4 & 5\\
         \midrule
         ProtoPool & \multirow{4}{*}{CUB-200-2011} & 3 & 26 & 107 & 151 & 113\\
         ProtoTree & & 143 & 76 & 73 & 53 & 55\\
         ProtoPool w/o & & \multirow{2}{*}{100} & \multirow{2}{*}{61} & \multirow{2}{*}{88} & \multirow{2}{*}{98} & \multirow{2}{*}{53}  \\
         focal similarity& & & & & &\\
         \midrule
         ProtoPool & \multirow{4}{*}{Stanford Cars} & 12 & 57 & 139 & 116 & 76\\
         ProtoTree & & 21 & 101 & 106 & 108 & 64\\
         ProtoPool w/o  &  & \multirow{2}{*}{70} & \multirow{2}{*}{103} & \multirow{2}{*}{96} & \multirow{2}{*}{79} & \multirow{2}{*}{52} \\
         focal similarity & & & & & &\\
         \bottomrule
    \end{tabular}
    
    \label{tab:user}
\end{table}

\section{Limitations}
Our ProtoPool model inherits its limitations from the other prototype-based models, including non-obvious prototype meaning. Hence, even after prototype projection from a training dataset, there is still uncertainty on which attributes it represents. However, there exist ways to mitigate this limitation, e.g. using a framework defined in~\cite{nauta2020looks}. Additionally, the choice of Gumbel-Softmax temperature $\tau$ and its decreasing strategy are not straightforward and require a careful hyperparameter search. Lastly, in the case of ProtoPool, increasing the number of prototypes does not increase the model accuracy after some point because the model saturates.

\section{Negative impact}
We base our solution on prototypical parts, which are vulnerable to a new type of adversarial attacks~\cite{hoffmann2021looks}. Hence, practitioners must consider this danger when deploying a system with a ProtoPool. Additionally, it can spread disinformation when prototypes derive from spoiled data or are presented without an expert comment, especially in fields like medicine.

\section{Additional discussion}
\begin{figure}[h]
    \vspace{-1em}
    \centering
    \includegraphics[width=0.8\textwidth]{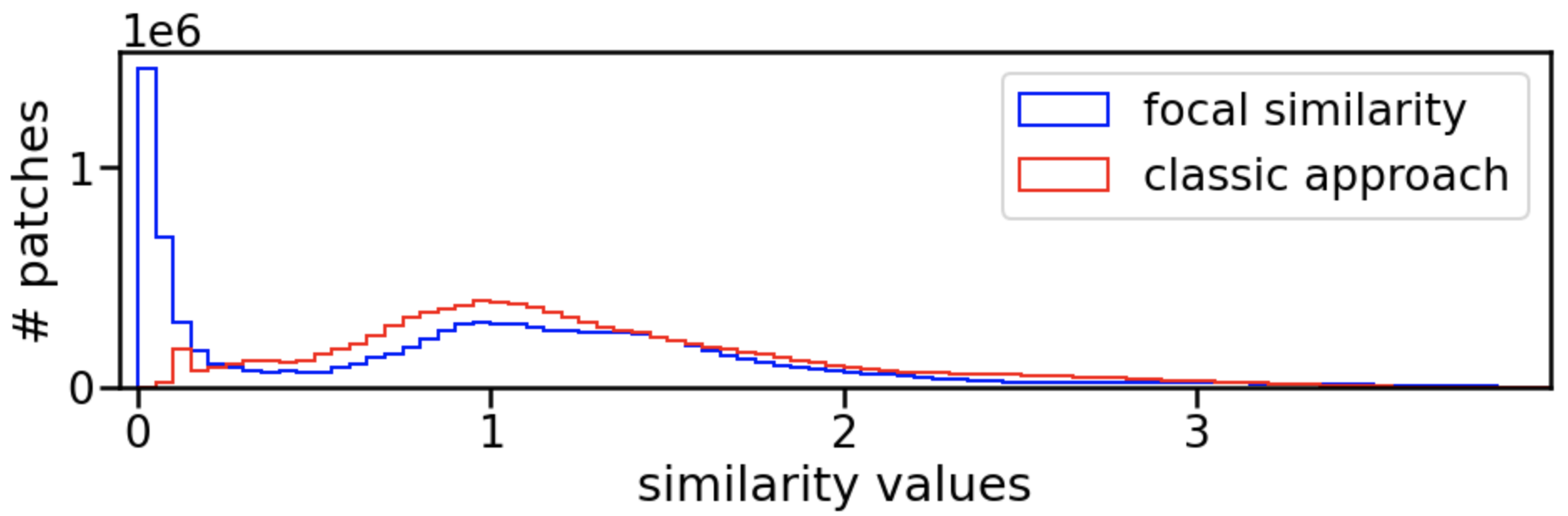}
    \caption{Distribution of similarity values for focal similarity and classic approach~\cite{rymarczyk2021protopshare} over $1000$ images. As a result of replacing the original approach with focal similarity, the distribution changes from unimodal to bimodal. }
    \label{fig:hist_sim}
\end{figure}

\paragraph{Why focal similarity works -- the intuition.} Focal similarity computes the similarity between patches and prototypes, which is then passed to the classification layer of the network, where standard CE loss is used. The big advantage of focal similarity is its ability to propagate gradient through all patches by subtracting the mean from the maximum similarity. In contrast to the original approach~\cite{rymarczyk2021protopshare}, which propagates gradient only through the patch with the maximum similarity. This way, ProtoPool generates salient prototypes that activate only in a few locations and return values close to zero for the remaining image parts (see Fig.~\ref{fig:hist_sim}). From this perspective, using the median instead of the mean would again limit gradient propagation to two patches (with maximum and median similarity).

\paragraph{Saturation of model capacity.}
The model reaches a plateau for around 200 prototypes, and there is no gain in further increase of prototype number. Therefore, the practitioners cannot sacrifice some interpretability to gain higher accuracy. In fact, this trend is also observed in the other methods with shared prototypes, like ProtoTree (see Fig.~7 in~\cite{nauta2021neural}). While we have no clear explanation for this phenomenon, we assume it can be caused by the entanglement of the prototypes. Therefore, one possible solution would be to enforce the prototypes orthogonality, as proposed in TesNet~\cite{wang2021interpretable}. The other option would be to modify the training procedure so that it iteratively adds new slots to each class corresponding to new pools of prototypes.
%

\paragraph{Focal similarity vs. reasoning type}
While the negative reasoning process draws conclusions based on the prototype's absence ("this does not look like that prototype"), the focal similarity concludes based on the prototype's presence ("this looks like that salient prototype, which usually occurs only one time"). For example, the negative reasoning could say: "this is a goat because it has no wings", while the focal similarity would rather say: "this is a goat because it has a goatee (a salient goat feature)".

\paragraph{Necessity for sharing prototypes}
We would like to recall the observation provided in~\cite{rymarczyk2021protopshare}. It shows that after training ProtoPNet with exclusive prototypes, patches' representations are clustered around prototypes of their true classes, and the prototypes from different classes are well-separated. As a result, the prototypes with similar semantics can be distant in representation space (see Fig. 2 in~\cite{rymarczyk2021protopshare}), resulting in unstable predictions. That is why it is essential to share the prototypes between classes.

\paragraph{User studies statistics} We provide additional statistics regarding the results of user studies. The confidence intervals are as follows: ProtoPool (our): $3.66\pm 2.00$, 
ProtoPool w/o focal similarity: $2.85\pm 2.63$, 
ProtoTree: $2.87\pm 2.68$. 
Moreover, we performed a Mann-Whitney U test to determine whether of ProtoPool (our) scores are higher than 'ProtoPool w/o focal similarity and ProtoTree. We obtained $p$-value$=2.78\cdot 10^{-12}$ and $1.16\cdot 10^{-11}$, respectively. Since both $p$-values are smaller than $0.05$, we reject the null hypothesis and conclude that the ProtoPool is significantly better than other methods.

\paragraph{Prototypes in a hard vs soft assignment} Through the development process, we experimented with the soft assignment (Softmax instead of Gumbel-Softmax). However, we observed that the model struggles to separate the prototypes in the latent space, even with the increased weight of a separation cost, and prototypes usually converge to one point in representation space. On the other hand, using only the hard assignments hinders the change in assignments during training. That is why we decided to use a hybrid approach, where we start with soft assignments and binarize them with Gumbel-Softmax. It allows the cluster and separation costs to roughly organize latent space with a soft assignment at the beginning and then refine it as the hard assignment dominates.



\end{document}